\documentclass[11pt,a4paper]{article}
\usepackage[hyperref]{emnlp2020}
\usepackage{times}
\usepackage{latexsym}
\usepackage{amsmath}
\usepackage{xspace}
\usepackage{booktabs}
\usepackage{multirow}
\usepackage{comment}
\usepackage{amssymb}
\usepackage{amsfonts}
\usepackage{enumerate}
\usepackage{caption}
\usepackage{subcaption}
\usepackage[T1]{fontenc}
\usepackage[utf8]{inputenc}

\usepackage{ltablex}
\usepackage{caption, booktabs}
\usepackage{cuted}
\usepackage[T5,T1]{fontenc}
\usepackage{combelow}
\usepackage[normalem]{ulem}

\DeclareTextSymbolDefault{\ohorn}{T5}
\DeclareTextSymbolDefault{\uhorn}{T5}

\usepackage{microtype}
\usepackage{cleveref}
\crefname{section}{\S}{\S\S}
\Crefname{section}{\S}{\S\S}
\crefname{table}{Tab.}{}
\crefname{figure}{Fig.}{}
\crefname{algorithm}{Algorithm}{}
\crefname{equation}{eq.}{}
\crefname{appendix}{App.}{}
\crefname{thm}{Theorem}{}
\crefname{prop}{Proposition}{}
\crefname{cor}{Corollary}{}
\crefname{observation}{Observation}{}
\crefname{assumption}{Assumption}{}
\crefformat{section}{\S#2#1#3}

\usepackage{bbm}
\usepackage[disable]{todonotes}
\makeatletter
\newcommand*\iftodonotes{\if@todonotes@disabled\expandafter\@secondoftwo\else\expandafter\@firstoftwo\fi}  %
\makeatother
\newcommand{\noindentaftertodo}{\iftodonotes{\noindent}{}}
\newcommand{\note}[4][]{\todo[author=#2,color=#3,size=\scriptsize,fancyline,caption={},#1]{#4}} %

\newcommand{\notewho}[3][]{\note[#1]{#2}{blue!40}{#3}}     %

\newcommand{\ryan}[2][]{\note[#1]{ryan}{violet!40}{#2}}

\newcommand{\lucas}[2][]{\note[#1]{lucas}{blue!10}{#2}}
\newcommand{\Lucas}[2][]{\lucas[inline,#1]{#2}\noindentaftertodo}
\newcommand{\adina}[2][]{\note[#1]{adina}{green!10}{#2}}

\newcommand{\vh}{{\boldsymbol h}}
\newcommand{\vhC}{{\boldsymbol h}_C}
\newcommand{\vx}{\mathbf{x}}

\newcommand{\vhi}[1]{{\vh}^{(#1)}}

\newcommand{\vm}{\mathbf{m}}

\newcommand{\vmu}{{\boldsymbol \mu}}
\newcommand{\vmuv}{\vmu_v}
\newcommand{\Sigmav}{\Sigma_v}
\newcommand{\R}{\mathbb{R}}
\newcommand{\calA}{\mathcal{A}}
\newcommand{\calV}{\mathcal{V}}
\newcommand{\calN}{\mathcal{N}}
\newcommand{\calD}{\mathcal{D}}
\newcommand{\calDv}{\mathcal{D}^{(v)}}
\newcommand{\calVa}{{\calV}_a}
\newcommand{\calO}{\mathcal{O}}

\newcommand{\att}[1]{\textsc{#1}}
\newcommand{\MI}{\mathrm{MI}}
\newcommand{\NIW}{\mathrm{GIW}}
\newcommand{\IW}{\mathrm{IW}}
\newcommand{\LBNMI}{LBNMI\xspace}
\newcommand{\LBMI}{LBMI\xspace}
\newcommand{\LBA}{LBA\xspace}

\newcommand{\ent}{\mathrm{H}}

\newcommand{\Tr}{\mathrm{Tr}}
\newcommand{\Nv}{N_v}
\newcommand{\vtheta}{{\boldsymbol \theta}}
\newcommand{\vthetav}{\vtheta_v}

\newcommand{\pthetaC}{p_{\vtheta_C}}

\newcommand{\NPHard}{{NP-Hard}\xspace}

\newcommand{\defn}[1]{\textbf{#1}}
\newcommand{\bert}{BERT\xspace}
\newcommand{\elmo}{ELMo\xspace}

\newcommand{\XX}{36\xspace}

\newcommand{\unimorph}{UniMorph\xspace}
\newcommand{\fasttext}{fastText\xspace}
\newcommand{\linspector}{LIN\-SPECTOR\xspace}

\newcommand{\truep}{\overline{p}}

\newcommand{\breakslash}{/\allowbreak\xspace}

\definecolor{bertcolorname}{rgb}{0.4, 0.76, 0.65}
\definecolor{fasttextcolorname}{rgb}{0.99, 0.55, 0.3}

\definecolor{scattercolorA}{rgb}{0.851,0.373,0.008}
\definecolor{scattercolorB}{rgb}{0.106,0.619,0.467}

\newcommand{\bertC}{\textcolor{bertcolorname}{\bert}\xspace}
\newcommand{\fasttextC}{\textcolor{fasttextcolorname}{\fasttext}\xspace}

\newcommand{\scatcolA}[1]{\textcolor{scattercolorA}{#1}}
\newcommand{\scatcolB}[1]{\textcolor{scattercolorB}{#1}}

\DeclareMathOperator*{\argmax}{argmax}

\newcommand{\mathcheck}[1]{#1}
\newcommand{\textcheck}[1]{#1}
\everypar{\looseness=-1}

\aclfinalcopy %
\def\aclpaperid{2122} %

\title{Intrinsic Probing through Dimension Selection}

\usepackage{emoji}
\newcommand{\ucambridge}{\emoji[twitter]{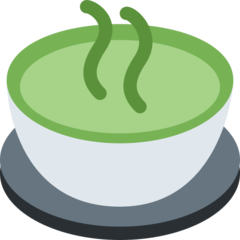}}
\newcommand{\ethz}{\emoji[twitter]{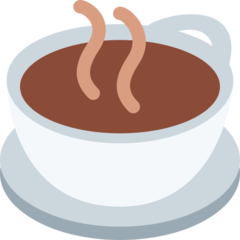}}
\newcommand{\fairesearch}{\emoji[twitter]{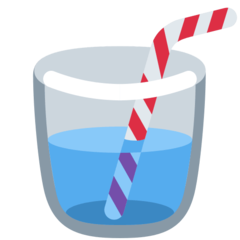}}
\newcommand{\mila}{\emoji[twitter]{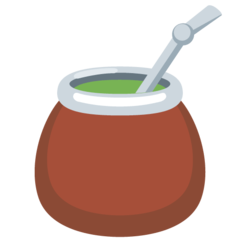}}

\author{Lucas Torroba Hennigen$^{\mila}$~\;~\;~Adina Williams$^{\fairesearch}$~\;~\;~Ryan Cotterell\,$^{\ucambridge,\ethz}$ \\
  $^{\mila}$Qu{\'e}bec Artificial Intelligence Institute (Mila)~\;~\;~$^{\ucambridge}$\,University of Cambridge \\
  $^{\fairesearch}$Facebook AI Research~\;~\;~$^{\ethz}$\,ETH Z\"{u}rich \\
  \texttt{lucas.torroba-hennigen@mila.quebec},~\;~\;~ \texttt{adinawilliams@fb.com},\\
  \texttt{ryan.cotterell@inf.ethz.ch}
}

\date{}

\begin{document}
\maketitle
\begin{abstract}

Most modern NLP systems make use of pre-trained contextual representations that attain astonishingly high performance on a variety of tasks.
Such high performance should not be possible unless some form of linguistic structure inheres in these representations, and a wealth of research has sprung up on probing for it.
In this paper, we draw a distinction between intrinsic probing, which examines \emph{how linguistic information is structured} within a representation, and the extrinsic probing popular in prior work, which only argues for the \emph{presence} of such information by showing that it can be successfully extracted.
To enable intrinsic probing, we propose a novel framework based on a decomposable multivariate Gaussian probe that allows us to determine whether the linguistic information in word embeddings is dispersed or focal. We then probe \fasttext and \bert for various morphosyntactic attributes across \XX languages.
\textcheck{We find that most attributes are reliably encoded by only a few neurons, with \fasttext concentrating its linguistic structure more than \bert.}\footnote{Code and data are available at \url{https://github.com/rycolab/intrinsic-probing}.}
\end{abstract}

\section{Introduction}

Natural language processing (NLP) is enamored of contextual word representations---and for good reason! Contextual word-embedders, e.g. 
\bert~\cite{devlin-etal-2019-bert} and \elmo~\cite{petersDeepContextualizedWord2018},
have bolstered NLP
model performance on myriad tasks, such as syntactic parsing
\citep{kitaev-etal-2019-multilingual}, coreference resolution \citep{joshi-etal-2019-bert}, morphological tagging \citep{kondratyuk-2019-cross} and text generation \citep{zellersHellaSwagCanMachine2019}. Given the large empirical gains observed when they are employed, it is all but certain that word representations derived from neural networks \textcheck{encode some continuous analogue of linguistic structures.}\lucas{Took off the next sentence, it was redundant}

Exactly \emph{what} these representations encode about linguistic structure, however, remains little understood.
Researchers have studied this question by attributing function to specific network cells with visualization methods~\citep{karpathyVisualizingUnderstandingRecurrent2015,liVisualizingUnderstandingNeural2016} and by \defn{probing}~\citep{alain, belinkov-probing}, which seeks to extract structure from the representations.
Recent work has probed various representations
for correlates of morphological \cite{belinkov-etal-2017-neural, giulianelliHoodUsingDiagnostic2018}, syntactic \cite{hupkes2018visualisation,zhang-bowman-2018-language,hewitt-manning-2019-structural,lin-etal-2019-open}, and semantic \cite{kim-etal-2019-probing} structure.

  \begin{figure}[]
    \centering
    \includegraphics[width=\linewidth]{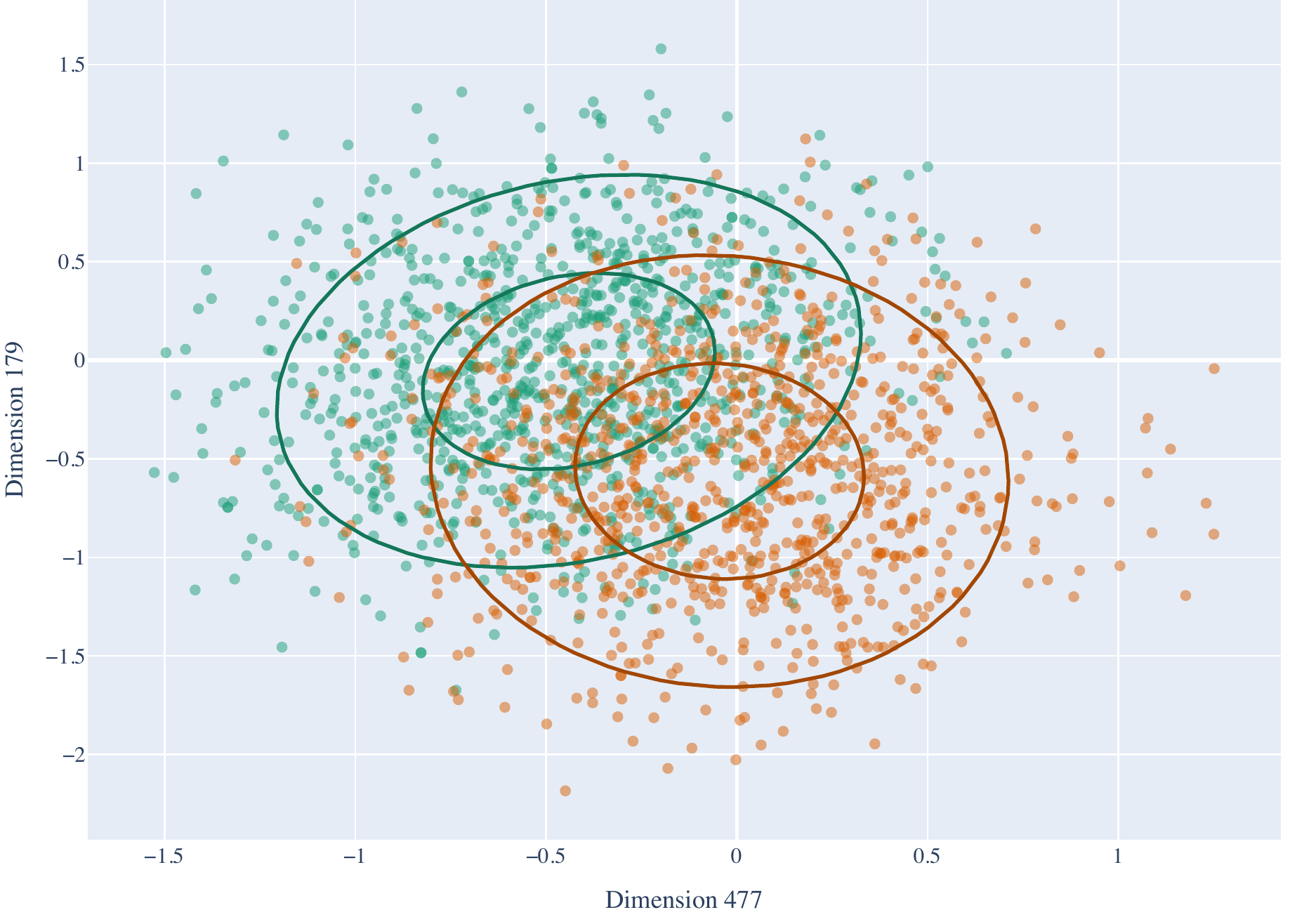}
      \caption{Scatter plot of the two most informative \bert dimensions for English \scatcolA{present} and \scatcolB{past} tense.
      The contours belong to our probe.
      }
      \label{fig:first-page}
  \end{figure}

Most current probing efforts focus on what we term \defn{extrinsic probing}, where the goal is to determine whether the posited linguistic structure is predictable from the learned representation. 
Generally, extrinsic probing works argue for the presence of linguistic structure by showing that it is extractable from the representations using a machine learning model.
In contrast, we focus on \defn{intrinsic probing}---whose goals are a proper superset of the goals of extrinsic probing. In intrinsic probing, one seeks to determine not only whether a signature of linguistic structure can be found, but also how it is encoded in the representations. 
In short, we aim to discover which particular  ``neurons'' (a.k.a.\, dimensions) in the representations \emph{correlate} with a given linguistic structure.\ryan{Hang!}\lucas{Which sentence does this refer to? Also, Hang == delete?}
Intrinsic probing also has ancillary benefits that extrinsic probing lacks; it can facilitate manual analyses of representations and potentially yield a nuanced view about the information encoded by them.

The technical portion of our paper focuses on developing a novel framework for intrinsic probing: we scan sets of dimensions, or neurons, in a word vector representation which activate when they correlate with target linguistic properties. 
We show that when intrinsically probing high-dimensional representations, the present \textcheck{probing} paradigm
\ryan{What is the present paradigm? I am confused.}\lucas{better?}
is insufficient (\cref{sec:probing-dimension-selection}).
Current probes are too slow \textcheck{to be used under our framework}, which invariably leads to low-resolution scans that can only look at one or a few neurons at a time.\ryan{The logical structure of the previous two paragraphs isn't tight enough for my liking.}\lucas{Is this better?}
Instead, we introduce \defn{decomposable probes}, which can be trained once on the whole representation and henceforth be used to scan \emph{any} selection of neurons.
To that end, we describe one such probe that leverages the multivariate Gaussian distribution's inherent decomposability, and evaluate its performance on a large-scale, multi-lingual, morphosyntactic probing task (\cref{sec:model}).

We experiment on \XX languages\footnote{See \cref{sec:probed-attributes} for a list.} from the Universal Dependencies treebanks~\citep{ud-2.1}.
We find that all the morphosyntactic features we considered are encoded by a relatively small selection of neurons. 
In some cases, very few neurons are needed; for instance, for multilingual \bert English representations, we see that, with two neurons, we can largely separate past and present tense in \cref{fig:first-page}.
In this, our work is closest to \newcite{lakretzEmergenceNumberSyntax2019}, except that we extend the investigation beyond \emph{individual} neurons---a move which is only made tractable by decomposable probing.
We also provide analyses on morphological features beyond number and tense.
Across all languages, 35 \textcheck{out of 768 neurons} on average suffice to reach a reasonable amount of encoded information, and adding more yields diminishing returns (see \cref{fig:representation-information}). Interestingly, in our head-to-head comparison of \bert and \fasttext, we find
that \fasttext almost always encodes information about morphosyntactic properties using fewer dimensions.

\section{Probing through Dimension Selection}
\label{sec:probing-dimension-selection}
The goal of intrinsic probing is to reveal how ``knowledge'' of a target linguistic property is structured within a neural network-derived representation. 
\textcheck{If said property can be predicted from the representations, we expect that this is because the neural network encodes this property~\citep{giulianelliHoodUsingDiagnostic2018}.}\ryan{We should get rid of the bigram ``demonstrates knowledge''. I would prefer ``predictable from'' or something that is closer to what we actually do}\lucas{better?}
We can then determine whether a probe requires a large subset or a small subset of dimensions to predict the target property reliably.\footnote{By analogy to the ``distributed'' and ``focal'' neural processes in cognitive neuroscience (see e.g.\, \citealt{bouton2018focal}), an intrinsic framework also imparts us with the ability to formulate much higher granularity hypotheses about whether particular morphosyntactic attributes will be widely or focally encoded in representations. }
Particularly small subsets could be used to manually analyze a network and its decision process, and potentially reveal something about how specific neural architectures learn to encode linguistic information.

To formally describe our framework, we first define the necessary notation.
We consider the probing of a word representation \mathcheck{$\vh \in \R^d$} for morphosyntax. In this work, our goal is find a subset of dimensions
\mathcheck{$C \subseteq D =  \{1, \ldots, d\}$} such that the corresponding subvector 
of \mathcheck{$\vhC$} contains only the dimensions that are necessary to predict the target morphosyntactic property we are probing for. %
For all possible subsets of dimensions \mathcheck{$C \subseteq D$}, and some random variable \mathcheck{$\Pi$} that
 ranges over \mathcheck{$P$} property values \mathcheck{$\{\pi_1, \ldots, \pi_P\}$}, we consider a general probabilistic probe:  \mathcheck{$\pthetaC(\Pi = \pi \mid \vhC)$}; note that the model is conditioned on \mathcheck{$\vhC$}, not on \mathcheck{$\vh$}. Our goal is to select a subset of dimensions using the log-likelihood of held-out data. We
 term this type of probing \defn{dimension selection}.
 One can express dimension selection as the following combinatorial optimization problem:
 \mathcheck{%
\begin{equation}\label{eq:optimization}
    C^\star = \argmax_{\substack{C \subseteq D, \\ |C| \leq k}} \sum_{n=1}^N \log \pthetaC(\pi^{(n)} \mid \vh^{(n)}_C)
\end{equation}
}
\textcheck{where \mathcheck{$\{ (\vhC^{(n)}, \pi^{(n)}) \}_{n=1}^N$} is a held-out dataset}.
Importantly, for complicated models we will require a \emph{different} parameter set \mathcheck{$\vtheta_C$} for each subset \mathcheck{$C \subseteq D$}. 
In the general case, solving a subset selection problem such as \cref{eq:optimization} is \NPHard~\citep{binshtokComputingOptimalSubsets2007}.
Indeed, without knowing more about the structure of \mathcheck{$\pthetaC$} we would have to rely on enumeration to solve this problem exactly.
As there are \mathcheck{${d \choose k}$} possible subsets, it  takes a prohibitively long
time to enumerate them all for even small \mathcheck{$d$} and \mathcheck{$k$}.

\paragraph{Greed is not Enough.}
A natural first approach to approximate a solution to \cref{eq:optimization}
is a greedy algorithm~\citep[Chapter 4]{kleinbergAlgorithmDesign2005}.
Such an algorithm chooses the dimension that results in the largest increase to the objective at every iteration. 
However, some probes, such as neural network probes, need to be trained with a gradient-based method for many epochs. 
In such a case, even a greedy approximation is prohibitively expensive. 
For example, to select the first dimension, we train \mathcheck{$d$} probes and take the best. 
To select the second dimension, we train \mathcheck{$d-1$} probes and take the best. 
This requires training \mathcheck{$\calO(dk)$} networks!
In the case of \bert, we have \mathcheck{$d=768$} and we would generally like to consider \mathcheck{$k$} at least up to \mathcheck{$50$}. 
Training on the order of \mathcheck{$38400$} neural models to probe for just \emph{one} morphosyntactic property is generally not practical. 
What we require is a \defn{decomposable} probe, which can be trained once on all dimensions and then be used to evaluate the log-likelihood of any subset of dimensions in constant or near-constant time.
To the best of our knowledge, no probes in the literature exhibit this property;
the primary technical contribution of the paper is the development of such a probe in \cref{sec:model}.

\paragraph{Other Selection Criteria.}
Our exposition above uses the log-likelihood of held-out data as a selection criterion for a subset of dimensions; however, any function that scores a subset of dimensions is suitable.
For example, much of the current probing literature relies on accuracy to evaluate probes~\citep[][\textit{inter alia}]{conneauWhatYouCan2018,liuLinguisticKnowledgeTransferability2019}, and two recent papers motivate a probabilistic evaluation with information theory~\citep{pimentelInformationTheoreticProbingLinguistic2020,voitaInformationTheoreticProbingMinimum2020}.
One could select based on accuracy, mutual information, or anything else within our framework.
In fact, recent work in intrinsic probing by \citet{dalviWhatOneGrain2019} could be recast into our framework if we chose a dimension selection criterion based on the magnitude of the weights of a linear probe.
However, we suspect that a performance-based dimension selection criterion (e.g., log-likelihood) should be 
more robust given that a weight-based approach is sensitive to feature collinearity, variance and regularization.
\textcheck{As we mentioned before, performance-based selection requires a probe to be decomposable, and to the best of our knowledge, this is not the case for the the linear probe of \newcite{dalviWhatOneGrain2019}.}
\ryan{This last sentence does not make sense to me!}\lucas{better?}

\section{A Decomposable Probe for Morphosyntactic Properties}\label{sec:model}

Using the framework introduced above, our goal is to probe for morphosyntactic properties in word representations.
We first describe the multivariate Gaussian distribution as it is responsible for our probe's decomposability~(\cref{sec:gaussian}),
and provide some more notation~(\cref{sec:probe-notation}).
We then describe our model~(\cref{sec:probe-description}) and a Bayesian formulation~(\cref{sec:bayesian}).

\subsection{Properties of the Gaussian}
\label{sec:gaussian}

The multivariate Gaussian distribution is defined as
\lucas{check definition in \url{http://www.inf.ed.ac.uk/teaching/courses/mlpr/2017/notes/w7c_gaussian_processes.html}}
\mathcheck{%
\begin{align}
    \calN&(\vx \mid \vmu, \Sigma)= \\
    & |2 \pi \Sigma|^{-\frac{1}{2}} \exp \left({ - \frac{1}{2} (\vx - \vmu)^\top \Sigma^{-1} (\vx - \vmu) }\right)\nonumber 
\end{align}
}
where \mathcheck{$\vmu$} is the mean of the distribution and \mathcheck{$\Sigma$} is the covariance matrix. We review the multivariate Gaussian with emphasis on the properties that make it ideal for intrinsic morphosyntactic probing. %

Firstly, it is decomposable.
Given a multivariate Gaussian distribution over \mathcheck{$\vx = [\vx_1 \,\, \vx_2]^\top$}
\lucas{check definition in \url{http://www.inf.ed.ac.uk/teaching/courses/mlpr/2017/notes/w7c_gaussian_processes.html}}
\mathcheck{%
\begin{align}
    p(\vx) &= \calN (\vx \mid \vmu, \Sigma) = \\ \nonumber
    &\calN \left(
\begin{bmatrix}
    \vx_1 \\ \vx_2
\end{bmatrix} \mathrel{\Big|}
\begin{bmatrix} \vmu_1 \\ \vmu_2 \end{bmatrix},
\begin{bmatrix}
    \Sigma_{11} & \Sigma_{12} \\
    \Sigma_{12}^\top & \Sigma_{22} 
\end{bmatrix}
\right)
\end{align}}
the marginals for $\vx_1$ and $\vx_2$ may be
 computed as
\lucas{Murphy 4.68}
\mathcheck{%
\begin{align}
    p(\vx_1) &= \calN(\vx_1 \mid \vmu_1, \Sigma_{11}) \\
    p(\vx_2) &= \calN(\vx_2 \mid \vmu_2, \Sigma_{22})
\end{align}
}
This means that if we know \mathcheck{$\vmu$} and \mathcheck{$\Sigma$}, we can obtain the parameters for \emph{any subset} of dimensions of \mathcheck{$\vx$} by selecting the appropriate subvector (and submatrix) of \mathcheck{$\vmu$~($\Sigma$)}.\footnote{The other variable, \mathcheck{$\Sigma_{12}$}, is a matrix that contains the covariances of each dimension of \mathcheck{$\vx_1$} with each dimension of \mathcheck{$\vx_2$}. We do not need it for our purposes.}
As we will see in \cref{sec:probe-description}, this property is the very centerpiece of our probe.
\mathcheck{%
Secondly, the Gaussian distribution is the maximum entropy distribution over the reals given a finite mean and covariance and no further information.}
Thus, barring additional information, the Gaussian is a good default choice.
\mathcheck{
\citet[Chapter 7]{jaynesProbabilityTheoryLogic2003} famously argued in favor of the Gaussian
because it is the real-valued distribution with support $(-\infty, \infty)$ that makes the fewest assumptions about the data (beyond its first two moments).}

\subsection{Notation for Morphosyntactic Probing}
\label{sec:probe-notation}
 
\Lucas{I reworked our explanation of the notation. Most importantly, I axed the sentence notation. Also added footnote and changed indexing.}

\textcheck{We now provide some notation for our morphosyntactic probe.
Let \mathcheck{$\{ \vh^{(1)}, \ldots, \vh^{(N)} \}$} be word representation vectors in \mathcheck{$\R^d$} for
\mathcheck{$N$} words \mathcheck{$\{ w^{(1)},\ldots,w^{(N)} \}$} from a corpus.
For example, these could be embeddings output by \fasttext~\citep{bojanowskiEnrichingWordVectors2017}, or contextual representations according to \elmo~\citep{petersDeepContextualizedWord2018} or \bert~\citep{devlin-etal-2019-bert}.
Furthermore, let \mathcheck{$\{ \vm^{(1)}, \ldots, \vm^{(N)}\}$}
be the morphosyntactic tags associated with each of those words in the sentential context in which they were found.\footnote{\textcheck{Crucially, some words may have different morphosyntactic tags depending on their context. For example, the number attribute of ``make'' could be either singular (``I make'') or plural (``They make'').}}}

Let \mathcheck{$\calA = \{a_1, \ldots, a_{|\calA|}\}$} be a universal\ryan{Cite the UniMorph paper here}\lucas{added it to footnote, is this what you meant? or do you want it in the main text?}\footnote{\textcheck{``Universal'' here refers to the set of all UniMorph dimensions and their possible features~\citep{sylak-glassmanCompositionUseUniversal2016,kirovUniMorphUniversalMorphology2018}}.} set of morphosyntactic attributes in a language, e.g. \att{person}, \att{tense}, \att{number}, etc. 
For each attribute \mathcheck{$a \in \calA$}, let \mathcheck{$\calVa$} be that attribute's universal set of possible values. For instance, we have \mathcheck{$\calV_{\att{person}} = \{1, 2, 3\}$} for most languages. 
For this task, we will
further decompose each morphosyntactic tag as a set of attribute--value pairs  \mathcheck{$\vm^{(i)} = \langle a_1\!=\!v_1, \ldots, a_{|\vm^{(i)}|}\!=\!v_{|\vm^{(i)}|}\rangle$}
where each attribute \mathcheck{$a_j$} is taken from the universal set of attributes
\mathcheck{$\calA$}, and each value \mathcheck{$v_j$} is taken from a set \mathcheck{$\calV_{a_j}$} of universal values specific to that attribute. For example, the morphosyntactic tag \mathcheck{$\vm$} for the English verb ``has'' would be \mathcheck{$\{\att{person} = 3, \att{number} = \textsc{sg}, \att{tense} = \textsc{prs} \}$}.

\subsection{Our Decomposable Generative Probe}
\label{sec:probe-description}

We now present our decomposable probabilistic probe.
We model the joint distribution between embeddings and a specific attribute's values
\mathcheck{%
\begin{equation}
p(\vh, v) = p(\vh \mid v) \, p(v)
\end{equation}}
where we define
\mathcheck{%
\begin{equation}
    p(\vh \mid v) = \calN(\vh \mid \vmuv, \Sigmav)
\end{equation}}
where \mathcheck{$\vmuv$} and \mathcheck{$\Sigmav$} are the value-specific mean and covariance.
We further define
\mathcheck{%
\begin{equation}
    p(v) = \mathrm{Categorical}\left(\calV_{a}\right)
\end{equation}}
\textcheck{This allows each value to have a different probability of occurring. This is important since our probe should be able to model that, e.g.\, the 3rd person is more prevalent than the 1st person in corpora derived from Wikipedia.}
\ryan{There must be a cleaner way to say this. Like, the Categorical has $|\calV_{a}|-1$ free parameters?}\lucas{I'd avoid parameters personally--I don't find them clear. is it clearer now?}
We can then probe with
\mathcheck{%
\begin{align}
    p(v \mid \vh) &= \frac{p(\vh, v)}{
      \sum_{v' \in \calVa} p(\vh, v')} \label{eq:conditional}
\end{align}}
which can be computed quickly as \mathcheck{$|\calVa|$} is small.\footnote{\unimorph's most varied attribute is \textsc{Case}, with 32 values, though most languages do not exhibit all of them.}
\mathcheck{This model is also known as quadratic discriminant analysis~\cite[Chapter 4]{murphyMachineLearningProbabilistic2012}.}\lucas{Check Murphy section 4.2.1}
\textcheck{Another interpretation of our model is that it} amounts to a generative classifier where, given some specific morphosyntactic attribute, we first sample one of its possible values \mathcheck{$v$}, and then sample an embedding from a value-specific Gaussian. 
Compared to a linear probe (e.g.\, \citealt{hewittDesigningInterpretingProbes2019})\lucas{I changed this from Hewitt and Manning to Hewitt and Liang. H\&M don't do classification, AFAIK}, whose decision boundary is linear for two values,
\mathcheck{the decision boundary of this model generalizes to conic sections, including parabolas, hyperbolas and ellipses~\citep[Chapter 4]{murphyMachineLearningProbabilistic2012}}.\lucas{Check Murphy Figure 4.3, and Wikipedia for QDA}

This formulation allows us to model the word representations of each attribute's value as a separate Gaussian.
Since the Gaussian distribution is decomposable~(\cref{sec:gaussian}), we can train a single model and from it obtain a probe for \emph{any} subset of dimensions in \mathcheck{$\calO(1)$} time.
To the best of our knowledge, no other probes in the literature possess this desirable property, which is what enables us to intrinsically probe representations for morphosyntax.

\subsection{Bayesically Done Now}
\label{sec:bayesian}

All that is left now is to obtain the value-specific Gaussian parameters \mathcheck{$\vthetav = \{\vmuv, \Sigmav\}$}.
Let \mathcheck{$\calDv = \{ \vhi{1}, \vhi{2}, \ldots, \vhi{\Nv} \}$} be a sample of
\mathcheck{$d$}-dimensional word representations for a value \mathcheck{$v$} for some language.
One simple approach is to use maximum-likelihood estimation (MLE) to estimate $\vthetav$; this amounts to computing the empirical mean and covariance matrix of $\calDv$.
However, \textcheck{in preliminary experiments we found that a Bayesian approach is advantageous since} it precludes degenerate Gaussians when there are more dimensions under consideration than training datapoints~\citep{srivastavaBayesianQuadraticDiscriminant2007}.\lucas{Murphy Section 4.2.5, bullet point 5.}

Under the Bayesian framework, we seek to compute the posterior distribution over the probe's parameters given our training data,
\mathcheck{%
\begin{align}
    p(\vthetav \mid \calDv) \propto p(\vthetav) \times p(\calDv \mid \vthetav)
\end{align}}
where \mathcheck{$p(\vthetav)$} is our Bayesian prior.
The prior encodes our \textcheck{\emph{a priori}} belief about the parameters in the absence of any data,
and \mathcheck{$p(\calDv \mid \vthetav)$} is the likelihood of the data under our model given a parameterization \mathcheck{$\vthetav$}.
\mathcheck{In the case of a Gaussian--inverse-Wishart prior,\footnote{Also known as a Normal--inverse-Wishart prior.}}\lucas{Murphy 4.6.3.2}
\mathcheck{%
\begin{align}
    p(\vthetav) &= \NIW(\vmuv, \Sigmav \mid \vmu_0, k_0, \Lambda_0, \nu_0) \\
     &= \calN(\vmuv \mid \vmu_0, \frac{1}{k_0}\Sigmav) \times \IW(\Sigmav \mid \Lambda_0, \nu_0) \nonumber
\end{align}}\lucas{Murphy Equation 4.201, relabelled.}
\textcheck{there is an exact expression for the posterior.}
\textcheck{The $\NIW$ prior has hyperparameters $\vmu_0, k_0, \Lambda_0, \nu_0$, where the inverse-Wishart distribution ($\IW$, see \cref{sec:inverse-wishart})
defines a distribution over covariance matrices~\citep[Chapter 4]{murphyMachineLearningProbabilistic2012}, and the Gaussian defines a distribution over the mean.}
\mathcheck{As this prior is conjugate to the multivariate Gaussian distribution,
our posterior over the parameters after observing
$\calD^{(v)}$ will also have a Gaussian--inverse-Wishart
distribution,
$\NIW(\vmuv, \Sigmav \mid \vmu_n, k_n, \Lambda_n, \nu_n)$,
with known parameters}~(see \cref{sec:niw-posterior-params}).\lucas{Murphy equation 4.209}

We did not perform full Bayesian inference as we found a maximum a posteriori (MAP) estimate to be sufficient for our purposes.\footnote{\mathcheck{The posterior predictive of this model is a Student's t-distribution~\citep{murphyConjugateBayesianAnalysis2007}.} Future work will explore a fully Bayesian implementation.}
MAP estimation uses the parameters at the posterior mode
\mathcheck{%
\begin{align}
    \vthetav^\star &= \argmax_{\vthetav} p(\vthetav \mid \calDv) \\
        &= \argmax_{\vmuv, \Sigmav} \NIW(\vmuv, \Sigmav \mid \vmu_n, k_n, \Lambda_n, \nu_n) \nonumber
\end{align}}
which are~\citep[Chapter 4]{murphyMachineLearningProbabilistic2012}
\mathcheck{%
\begin{align}
    \vmuv^\star &= \vmu_n \\
    \Sigmav^\star &= \frac{1}{\nu_n + d + 2} \Lambda_n
\end{align}
where $d$ is the dimensionality of the Gaussian.}\lucas{Murphy Section 4.6.3.4}

\section{Probing Metrics}\label{sec:metrics}

In this section, we describe the metrics that we compute.
We track both accuracy~(\cref{sec:accuracy}) and mutual information~(\cref{sec:mutual-information}).

\subsection{Accuracy}
\label{sec:accuracy}

As with most probes in the literature, we compute the accuracy of our model on held-out data.
We report the lower-bound accuracy (\LBA) of a set of dimensions \mathcheck{$C$}, which is defined as the highest accuracy achieved by any subset of dimensions \mathcheck{$C' \subseteq C$}.
This metric counteracts a decrease in performance due to the model overfitting in certain dimensions.
In principle, if a model was able to achieve a higher score using fewer dimensions, then there exists a model that can be at least as effective using a superset of those dimensions.

Despite its popularity, accuracy also has its downsides. 
In particular, we found it to be misleading when not taking a majority-class baseline into account, which complicates comparisons. For example, in \fasttext and \bert Latin (lat), our probe achieved slightly over 65\% accuracy when averaging over attributes. This appears to be high, but 65\% is the average majority-class baseline accuracy. Conversely, \LBNMI (see \cref{sec:mutual-information}) is roughly zero, which more intuitively reflects performance.
Hence, we prioritize mutual information in our analysis.

\subsection{Mutual Information}
\label{sec:mutual-information}

Recent work has advocated for information-theoretic metrics in probing~\citep{voitaInformationTheoreticProbingMinimum2020,pimentelInformationTheoreticProbingLinguistic2020}.
One such metric, mutual information (MI), measures how predictable the occurrence of one random variable is given another.

We estimate the MI between representations and particular attributes using
a method similar to the one proposed by \citet{pimentelMeaningFormMeasuring2019} (refer to \cref{sec:mutual-information-approx} for an extended derivation).
Let \mathcheck{$V_a$} be a \mathcheck{$\calVa$}-valued random variable denoting the value of a morphosyntactic attribute, and \mathcheck{$H$} be a \mathcheck{$\R^d$}-valued random variable for the word representation.

The mutual information between $V_a$ and $H$ is
\begin{equation}
\MI(V_a ; H) = \ent(V_a) - \ent(V_a \mid H)
\label{eq:mi}
\end{equation}
\textcheck{The attribute's entropy, \mathcheck{$\ent(V_a)$}, depends on the true distribution over values \mathcheck{$\truep(v)$}.
For this, we use the plug-in approximation \mathcheck{$p(v)$}, which is estimated from held-out data.}\lucas{Ryan, is this better?}
The conditional entropy, \mathcheck{$\ent(V_a \mid H)$} is trickier to compute, as it \textcheck{also} depends on the true distribution of embeddings given a value, \mathcheck{$\truep(\vh \mid v)$},
which is high-dimensional and poorly sampled in our data.\footnote{\mathcheck{When considering few dimensions in $\vh$ this can be estimated, e.g.\, by binning.} However, we cannot rely on such estimates for intrinsic probing in general.}
However, we can obtain an upper-bound if we use our probe \mathcheck{$p(v \mid \vh)$}  and compute~\citep{brownEstimateUpperBound1992}
\Lucas{CHECK THIS CAREFULLY. I changed the third line. This is slightly different from the appendix.}
\mathcheck{%
\begin{align}
  &\ent(V_a \mid H) \leq \ent_{p} (V_a \mid H) \label{eq:conditional-entropy-approx} \\
  &= -\sum_{v \in \calV_a} \truep(v) 
   \int \truep(\vh \mid v) \log_2 p(v \mid \vh)\,\mathrm{d} \vh \nonumber \\
  &\approx -\frac{1}{N} \sum_{n=1}^{N} \log_2 p(\tilde{v}^{(n)} \mid \tilde{\vh}^{(n)})
\end{align}
using held-out data, $\tilde{\calD} = \{(\tilde{\vh}^{(n)}, \tilde{v}^{(n)})\}_{n=1}^{N}$.}
\mathcheck{Incidentally, this is equivalent to computing the average negative log-likelihood of the probe on held-out data.}
Using these estimates in \cref{eq:mi}, we obtain \textcheck{an empirical} lower-bound on the MI.\notewho{tiago}{maybe you should say an empirical lower-bound, or an expected lower bound. Since you rely on the plug-in estimate being better than this, which is most likely true, but not proved-ly so.}\lucas{Fixed; left Tiago's comment for explanation.}

For ease of comparison across languages and morphosyntactic attributes, we define two metrics associated to MI.
The lower-bound MI (\LBMI) of any set of neurons \mathcheck{$C$} is defined as the \emph{highest} MI estimate obtained by any subset of those neurons \mathcheck{$C' \subseteq C$}.
While true MI can never decrease upon adding a variable, \textcheck{our estimate} may decrease due to overfitting in our model, or by it being unable to capture the complexity of \mathcheck{$\truep(\vh \mid v)$}.
\LBMI offers a way to counteract this limitation by using the very best estimate at our disposal for any set of dimensions.
In practice, we report lower-bound normalized MI (\LBNMI),
which normalizes \LBMI by the entropy of \mathcheck{$V_a$},
because normalizing MI estimates drawn from different samples enables them to be compared \citep{gates2019}.

\section{Experimental Setup}

In this section we outline our experimental setup.

\paragraph{Selection Criterion.}
We use log-likelihood as our greedy selection criterion.
We select 50 dimensions, and keep selecting even if the estimate has decreased.\footnote{Log-likelihood, unlike accuracy, is sensitive to confident but incorrect estimates. We found that this change allowed us to keep selecting dimensions that increase accuracy but decrease log-likelihood, as they may be informative but contain some noise or outliers.}

\paragraph{Data.}
We map the UD v2.1 treebanks~\citep{ud-2.1} to the \unimorph schema~\citep{kirovUniMorphUniversalMorphology2018,sylak-glassmanCompositionUseUniversal2016} using the mapping by~\citet{mccarthyMarryingUniversalDependencies2018}.
We keep only the ``main'' treebank for a language (e.g.\,UD\_Portuguese as opposed to UD\_Portuguese\_PUD).
We remove any sentences that would have a sub-token length greater than 512, the maximum allowed for our BERT model.\footnote{Out of a total of 419943 sentences in the treebanks, 4 were removed.}
We assign any tags from the constituents of a contraction to the contracted word form (e.g.,\ for Portuguese, we copy annotations from \textit{de} and \textit{a} to the contracted word form \textit{da}).
We use the UD train split to train a probe for each attribute, the validation split to choose which dimensions to select using our greedy scheme, and the test split to evaluate the performance of the probe after dimension selection.

We do not include in our estimates any morphological attribute--value pairs \textcheck{with fewer than 100 word types in \emph{any} of our splits, as we might not be able to model or evaluate them accurately.}
This removes certain constructions that mostly pertain to function words (e.g. as definiteness is marked only in articles in Portuguese, the attribute is dropped),
but we found it also removed rare inflected forms in our data, which may be due to inherent biases in the domain of text found in the treebanks (e.g. the future tense in Spanish).
\textcheck{We use all the words that have been tagged in one of the filtered attribute--value pairs (this includes both function and content words).}
Finally, we apply some minor post-processing to the annotations~(\cref{sec:ud-changes}).

\paragraph{Word Representations.}
We probe the multilingual \fasttext vectors,\footnote{We use the implementation by~\citet{graveLearningWordVectors2018}.} and the final layer of the multilingual release of \bert.\footnote{We use the implementation by~\citet{wolfHuggingFaceTransformersStateoftheart2020}.}
We compute word-level embeddings for \bert by averaging over sub-token representations as in~\citet{pimentelInformationTheoreticProbingLinguistic2020}.
We use the tokenization in the UD treebanks.

\paragraph{Hyperparameters.}
Our model has four hyperparameters, which control the Gaussian--inverse-Wishart prior.
We choose hyperparameter settings that have been shown to work well in the
literature~\citep{fraleyBayesianRegularizationNormal2007,murphyMachineLearningProbabilistic2012}.
\textcheck{We set $\vmu_0$ to the empirical mean, $\Lambda_0$ to a diagonalized version of the empirical covariance,
$\nu_0 = d + 2$, and $k_0 = 0.01$. We note that the resulting prior is degenerate if the data contains only one datapoint, since the covariance is not well-defined. However, since we do not consider attribute--values with less that 100 word types, this does not affect our experiments.}

\section{Results and Discussion}
\label{sec:results}

\begin{figure}[tb]
\centering
  \begin{subfigure}[]{\linewidth}
    \centering
    \includegraphics[width=\linewidth]{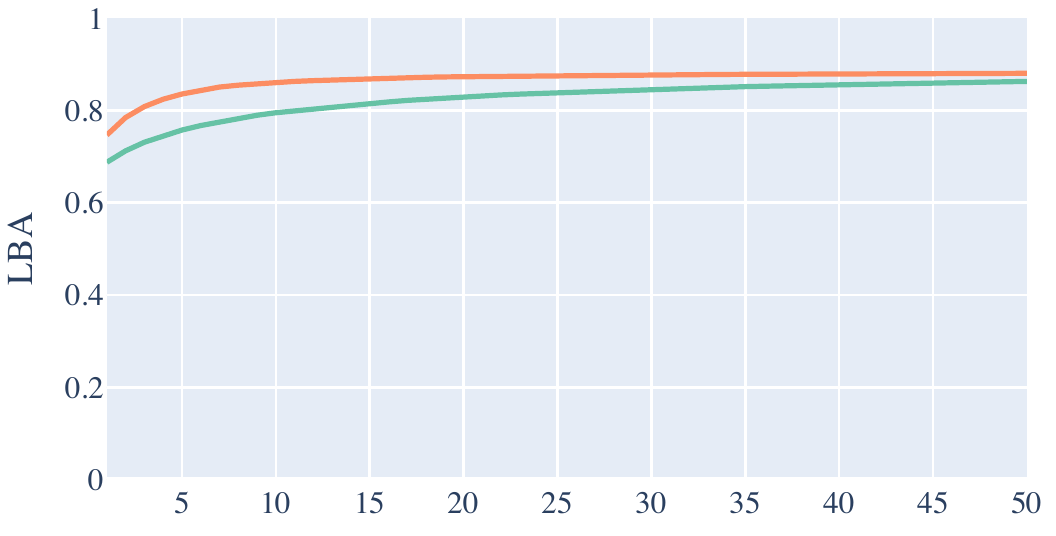}
  \end{subfigure}

  \begin{subfigure}[]{\linewidth}
    \centering
    \includegraphics[width=\linewidth]{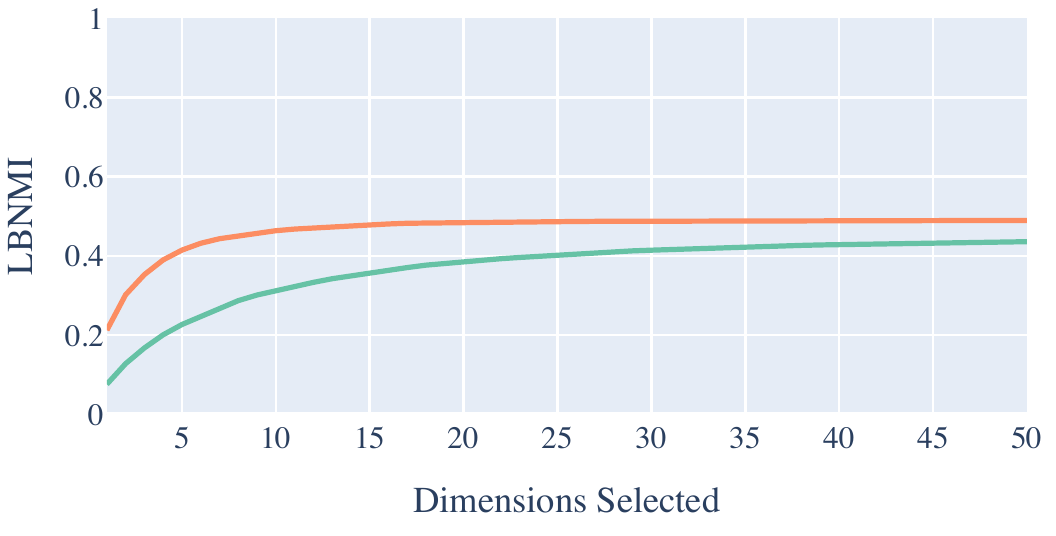}
  \end{subfigure}
    
\caption{
    The average lower-bound accuracy (\LBA) and lower-bound normalized mutual information (\LBNMI) across all evaluated attributes and languages for
    \fasttextC and \bertC.}
    \label{fig:representation-information}
\end{figure}

Overall, our results strongly suggest that morphosyntactic information tends to be highly focal (concentrated in a small set of dimensions) in \fasttext, whereas in \bert it is more dispersed.
Averaging across all languages and attributes~(\cref{fig:representation-information}),
\fasttext has on average $0.306$ \LBNMI at two dimensions, which is around twice as much as \bert at the same dimensionality.
However, the difference between the two becomes progressively smaller, reducing to $0.053$ at 50 dimensions.
A similar trend holds for \LBA~(\cref{sec:accuracy}),
with an even smaller difference at higher dimensions.
On the whole, roughly 10 dimensions are required to encode any morphosyntactic attribute we probed \fasttext for, compared to around 35 dimensions for \bert.

\begin{figure}[tb]
    \centering
      \includegraphics[type=pdf,ext=.pdf,read=.pdf,width=\linewidth]{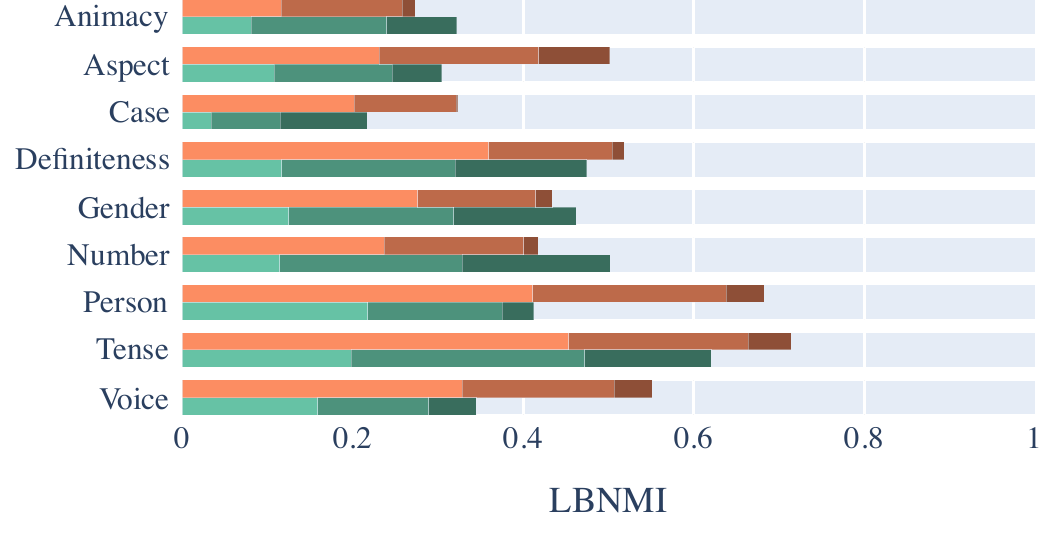}
    \caption{Comparison of per-attribute average lower-bound normalized mutual information (\LBNMI) for \fasttextC and \bertC.
        Each bar is broken up into three components, which denote the \LBNMI after selecting 2, 10 and 50 dimensions.
        }
    \label{fig:lbnmi-attrs}
\end{figure}

\begin{figure}[tbh]
    \centering
      \includegraphics[type=pdf,ext=.pdf,read=.pdf,width=\linewidth]{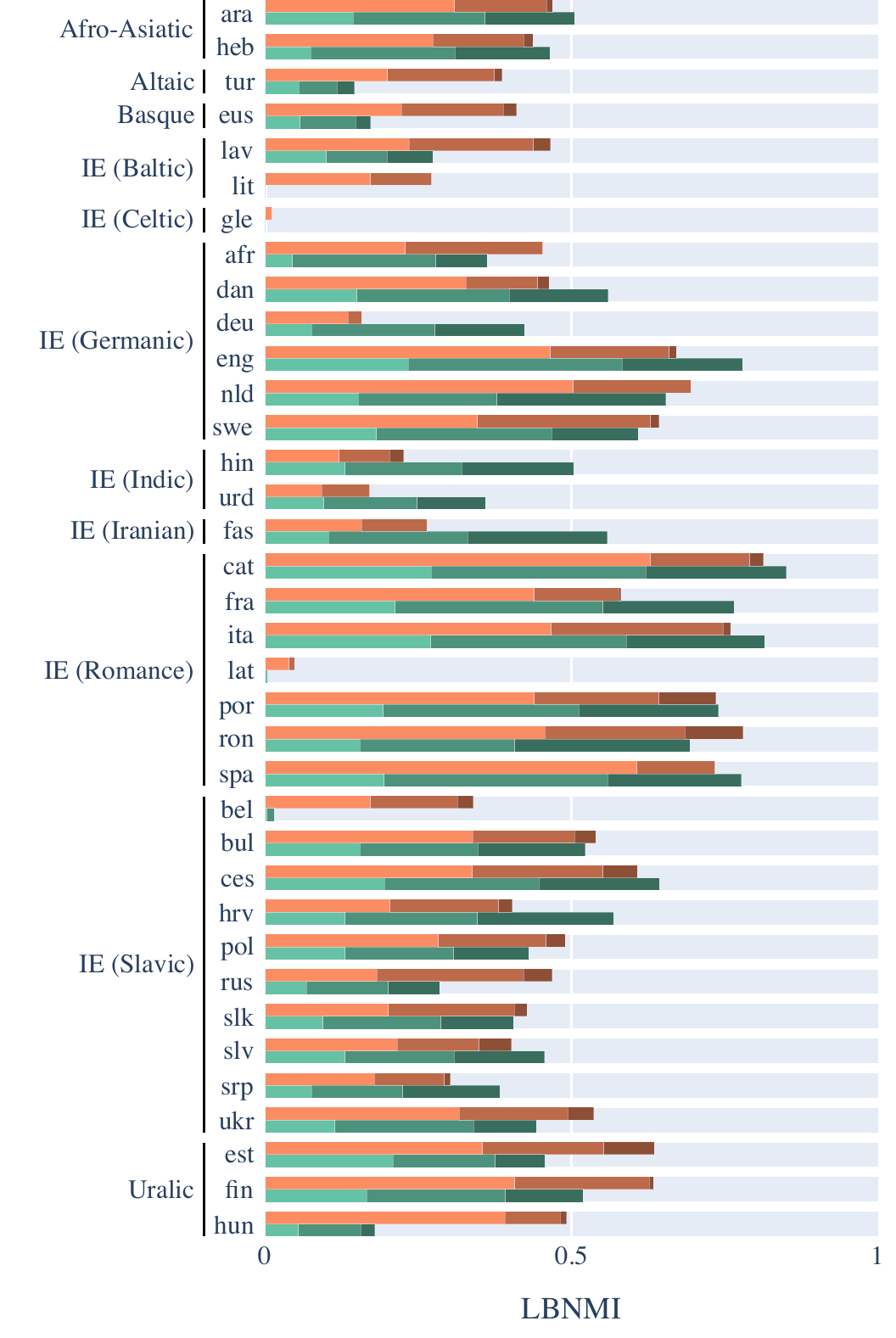}
    \caption{Comparison of per-language average lower-bound normalized mutual information (\LBNMI) for \fasttextC and \bertC.
        Each bar is broken up into three components, which denote the \LBNMI after selecting 2, 10 and 50 dimensions.
        }
    \label{fig:lbnmi-langs}
\end{figure}

The pattern above holds across attributes (\cref{fig:lbnmi-attrs}), and languages (\cref{fig:lbnmi-langs}).
There is little improvement in \fasttext performance when adding more than 10 dimensions and, in some cases, two \fasttext dimensions can explain half of the information achieved when selecting 50.
In contrast, while \bert also displays highly informative dimensions, a substantial increase in \LBNMI can be obtained by going from 2 selected dimensions to 10 and 50.
Among languages, the only exceptions to this are the Indic languages, where \bert concentrates more morphological information than \fasttext already at 2 dimensions.
Interestingly, when looking at attributes, our results suggest that \fasttext encodes most attributes better than \bert \textcheck{(when considering the 50 most informative dimensions)}, except animacy, gender and number. \adina{interesting, all hierarchically low noun features.}
These findings also hold for \LBA, where we additionally find little to no gain when comparing \LBA after 50 dimensions to accuracy on the full vector.

\begin{figure}[tb]
  \centering
  \begin{subfigure}[]{\linewidth}
    \centering
    \includegraphics[width=\linewidth]{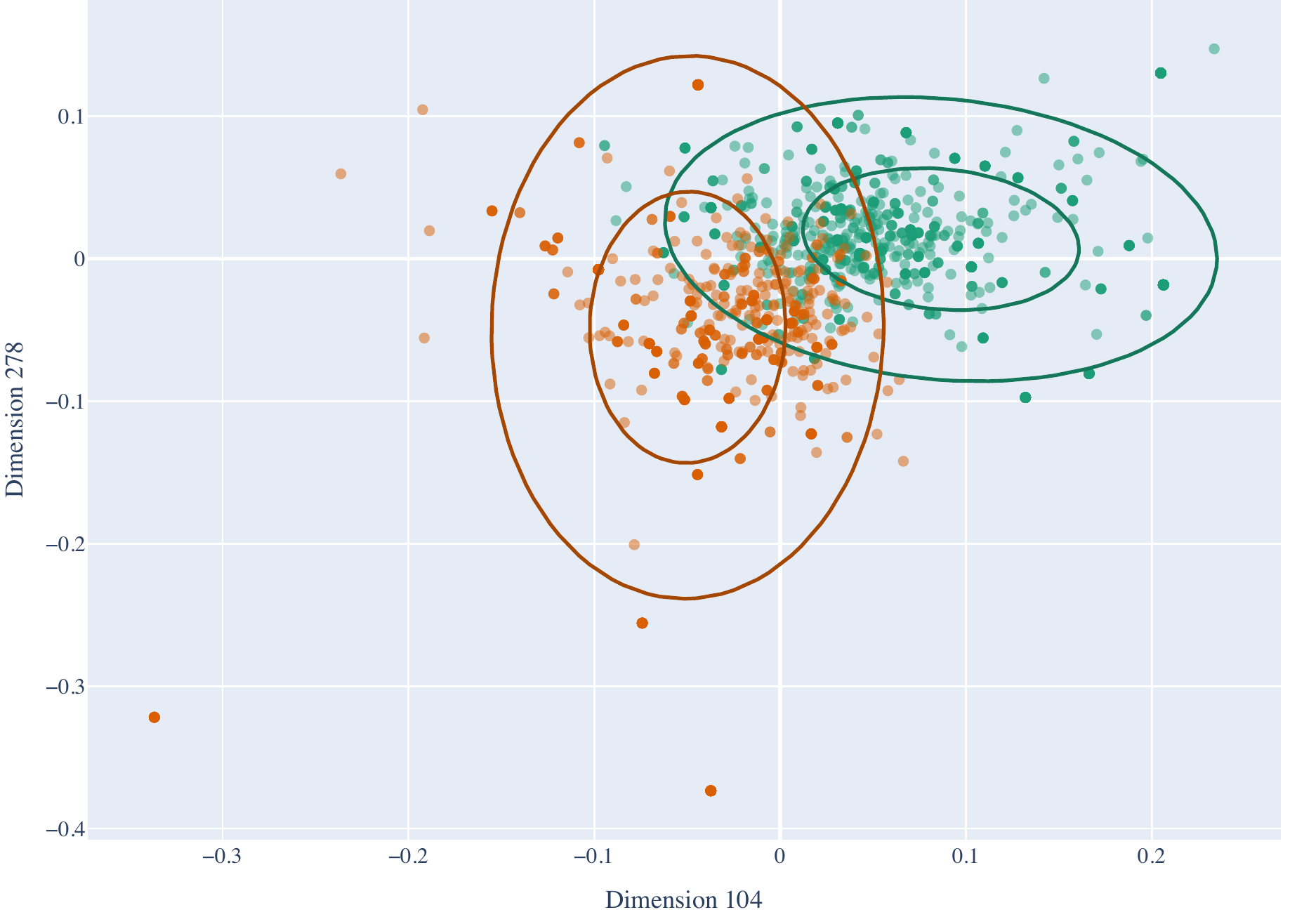}
  \end{subfigure}
  \\[1ex]

  \begin{subfigure}[]{\linewidth}
    \centering
    \includegraphics[width=\linewidth]{images/scatter_bert_eng_Tense.pdf}
  \end{subfigure}
  \caption{Scatter graph of two most informative \fasttext (above) and \bert (below) dimensions for English \scatcolA{present} and \scatcolB{past} tense.
  Contours belong to our probe.
  }
  \label{fig:eng-tense-scatter}
\end{figure}

Visualizing the most informative dimensions for \bert and \fasttext may give some intuition for how this trend manifests.
\cref{fig:eng-tense-scatter} shows a scatter plot of the two most informative dimensions selected by our probe for English tense in \fasttext and \bert.
We observed similar patterns for other morphosyntactic attributes.
Both embeddings have dimensions that induce some separability in English tense, but this is more pronounced in \fasttext than \bert.
We cannot clearly plot more than two dimensions at a time, but based on the trend depicted in \cref{fig:representation-information}, we can intuit that \bert makes up for at least part of the gap by inducing more separability as dimensions are added.

\begin{figure}[tb]
    \centering
    \includegraphics[width=\linewidth]{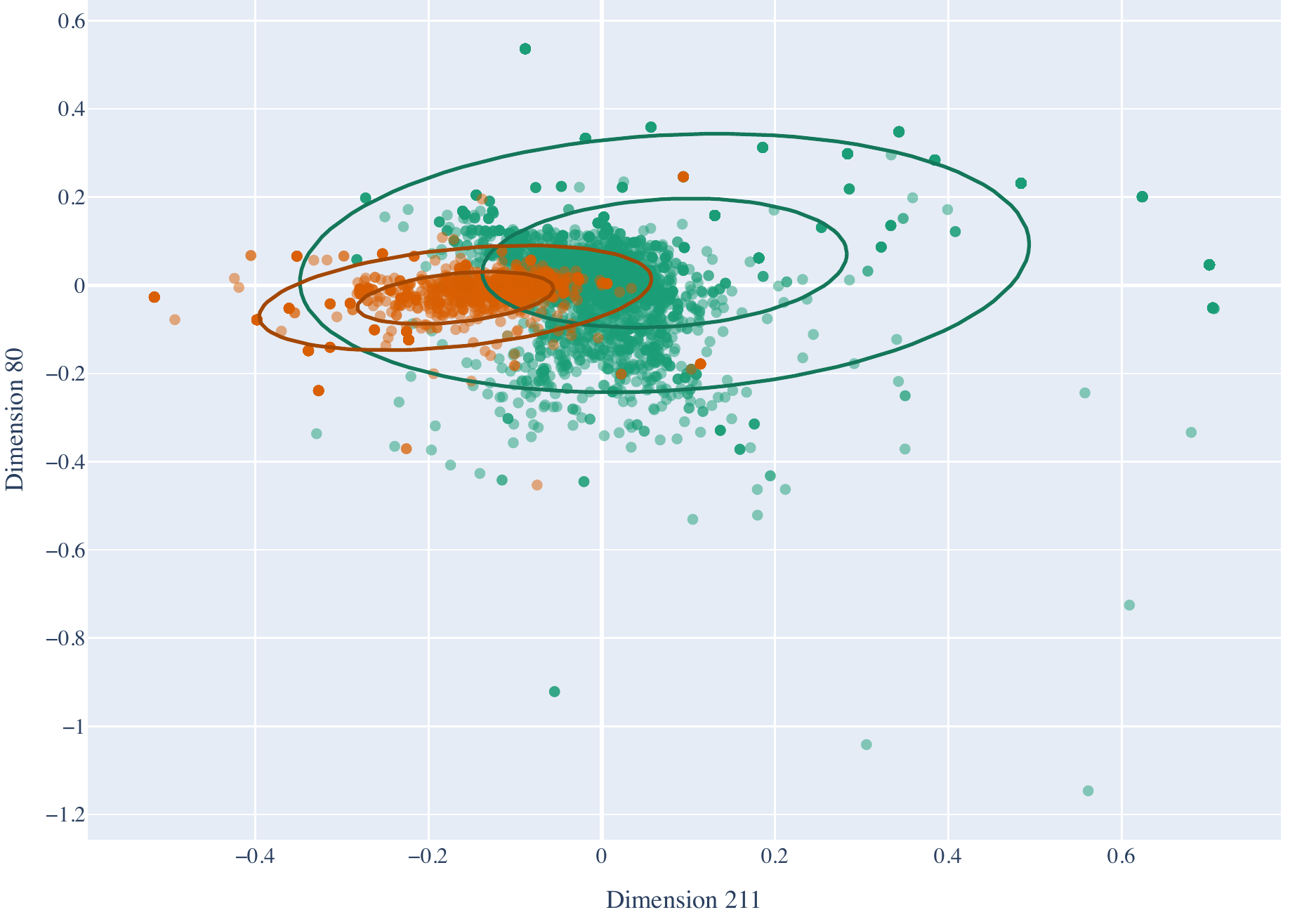}
    \caption{Two \fasttext dimensions that are informative for Portuguese number, but which do not appear jointly Gaussian. This dimensions pair was \emph{not} favored by our Gaussian probing model; we found it by modeling $p(\vh \mid v)$ with a Gaussian--Cauchy mixture model.}
    \label{fig:fasttext-non-gaussian}
\end{figure}

\begin{figure}[tb]
    \centering
    \includegraphics[width=\linewidth]{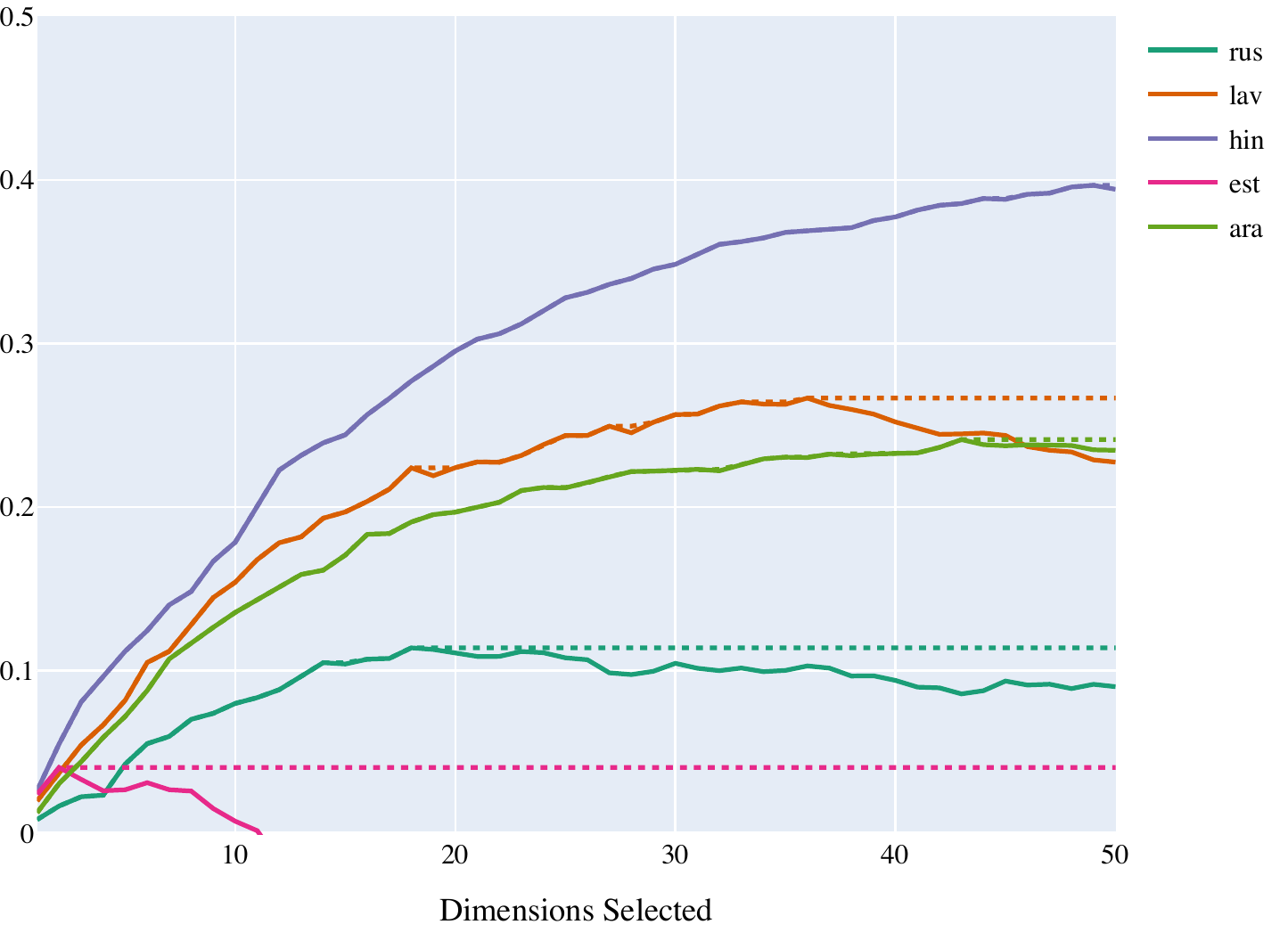}
    \caption{Plot of \LBNMI (\dotuline{dotted}) and normalized MI (\underline{solid}) curves for case in 5 randomly selected languages. Note that the $y$-axis ranges from $0$--$0.5$ unlike other graphs. Observe how the normalized MI estimates start to decrease after a certain number of dimension have been selected.}
    \label{fig:limitations}
\end{figure}

\textcheck{%
\subsection{Limitations}
\label{sec:limitations}
\todo{Check this section (and the associated Figures) with a bit more care if possible!}
The generative nature of our probe means that adequately modeling the embedding distribution \mathcheck{$p(\vh \mid v)$} is of paramount importance.
We choose a Gaussian model in order to assume as little as possible about the distribution of \bert and \fasttext embeddings; however, as one reviewer pointed out, the embedding distribution is unlikely to be Gaussian (see \cref{fig:fasttext-non-gaussian} for an example).
\ryan{Mention our empirical tests of this. We could do a komolgorov-smirnov test!}\lucas{we tried this, but this ended up being hard as tests are not designed for high-dimensional spaces like ours. I added a counterexample image.}
This results in a looser bound on the mutual information for dimensions in which the Gaussian assumption does not hold, which leads to decreasing mutual information estimates after a certain number of dimensions are selected (see \cref{fig:limitations}).
As we compute and report an empirical lower-bound on the mutual information for any subset of dimensions (\LBMI), we have evidence that there is \emph{at least} that amount of information for any given subset of dimensions.
However, we expect that better modeling of the embedding distribution should improve our bound on the mutual information and thus yield a better probe~\citep{pimentelInformationTheoreticProbingLinguistic2020}.
}

\section{Related Work}\label{sec:related-work}

\textcheck{%
There has been a growing interest in understanding what information is in NLP models' internal representations.
Studies vary widely, from detailed analyses of particular scenarios and linguistic phenomena~\citep{linzenAssessingAbilityLSTMs2016,gulordavaColorlessGreenRecurrent2018,ravfogelCanLSTMLearn2018, krasnowska-kierasEmpiricalLinguisticStudy2019,wallaceNLPModelsKnow2019,warstadt-etal-2019-investigating,
sorodocProbingReferentialInformation2020}
to extensive investigations across a wealth of tasks~\citep{tenneyWhatYouLearn2018,conneauWhatYouCan2018,liuLinguisticKnowledgeTransferability2019}.
A plethora of methods have been designed and applied~\citep[e.g.\,][]{liVisualizingUnderstandingNeural2016,saphraUnderstandingLearningDynamics2019,jumeletAnalysingNeuralLanguage2019a} to answer this question.
Probing~\citep{adiFinegrainedAnalysisSentence2017,hupkes2018visualisation,conneauWhatYouCan2018} is one prominent method, which consists of using a lightly parameterized model to predict linguistic phenomena from intermediate representations, albeit recent work has raised concerns on how model parameterization and evaluation metrics may affect the effectiveness of this approach~\citep{hewittDesigningInterpretingProbes2019,pimentelInformationTheoreticProbingLinguistic2020,maudslayTaleProbeParser2020,pimentelPareto}.}\ryan{We should cite our other lab's work in this section. We cite everybody but us. Both our ACL 2020 and other EMNLP 2020 papers!}\lucas{better?}

Most work in intrinsic probing has focused in the identification of individual neurons that
are important for a task~\citep{liVisualizingUnderstandingNeural2016, kadarFixed,liUnderstandingNeuralNetworks2017,lakretzEmergenceNumberSyntax2019}.
Similarly, \citet{clarkWhatDoesBERT2019} and \citet{voitaAnalyzingMultiHeadSelfAttention2019}
use probing to analyze BERT's attention heads, finding
some interpretable heads that attend to positional and syntactic features.
However, there has also been some work investigating collections of neurons.
For example, \citet{shiDoesStringBasedNeural2016} observe that different training objectives can affect how focal an intermediate representation is.
Recently, \citet{dalviWhatOneGrain2019} use the magnitude of the weights learned by a linear probe as a proxy for dimension informativeness, and find dispersion varies depending on linguistic category.
\citet{bauIdentifyingControllingImportant2019} use unsupervised methods to find neurons that are correlated across various models, quantify said correlation, and upon manual analysis find interpretable neurons.
\textcheck{%
In concurrent work in computer vision, \citet{bauUnderstandingRoleIndividual2020} identify units whose local, peak activations correlate with features in an image (e.g., material, door presence), show that ablation of these units has a disproportionately big impact on the classification of their respective features, and can be manually controlled, with interpretable effects.}

Most similar to our analysis is \linspector~\citep{sahinLINSPECTORMultilingualProbing2020}, a suite of probing tasks that includes probing for morphosyntax.
Our work differs in two respects.
Firstly, whereas \linspector focuses on extrinsic probing, we probe intrinsically.
Secondly, the scope of our morphosyntactic study is more typologically diverse (\XX vs. 5 languages), albeit they consider more varieties of word representations, such as GloVe~\citep{penningtonGloveGlobalVectors2014} and \elmo~\citep{petersDeepContextualizedWord2018}---but not \bert.

\section{Conclusion} \adina{the structure of this conclusion seems weird, what do you think \@ryan}\ryan{Agree should be re-written!}\lucas{Is this better?}

\textcheck{%
In this paper, we introduce an alternative framework for intrinsic probing, which we term dimension selection.
The idea is to use probe performance on different subsets of dimensions as a gauge for how much information about a linguistic property different subsets of dimensions jointly encode.
We show that current probes are unsuitable for intrinsic probing through dimension selection as they are not inherently decomposable, which is required to make the procedure computationally tractable.
Therefore, we present a decomposable probe which is based on the Gaussian distribution,} and evaluate its effectiveness by probing \bert and \fasttext for morphosyntax across \XX languages.
Overall, we find that \fasttext is more focal than \bert, requiring fewer dimensions to capture most of the information pertaining to a morphosyntactic property.

\textcheck{%
\paragraph{Future Work.}
Future work will be separated into two strands.
The first will focus on how to better model the distribution of embeddings given a morphosyntactic attribute; as mentioned above, this should yield a better probe overall.
The second strand of work pertains to a deeper analysis of our results, and expansion to other probing tasks.\lucas{modified last sentence}
}

\section*{Acknowledgments}
We thank Tiago Pimentel, Evgeny Kharitonov, Mark Tygert, Marco Baroni and the anonymous reviewers for their helpful comments.

\bibliography{lucas-refs,acl2020,anthology,misc}
\bibliographystyle{acl_natbib}

\newpage
\appendix

\section{Gaussian--inverse-Wishart Posterior Parameters}
\label{sec:niw-posterior-params}

Using the notation introduced in \cref{sec:bayesian}, \mathcheck{the parameters of the Gaussian--inverse-Wishart distribution
$\NIW(\vmuv, \Sigmav \mid \vmu_n, k_n, \Lambda_n, \nu_n)$, are~\citep{murphyMachineLearningProbabilistic2012}
\begin{align}
  \vmu_n &= \frac{k_0 \vmu_0 + \Nv \bar{\vh}}{k_n}\\
  k_n &= k_0 + \Nv \\
  \nu_n &= \nu_0 + \Nv \\
  \Lambda_n &= \Lambda_0 + S \\
    &\hspace{-0.5em} + \frac{N_v k_0}{\Nv + k_0}(\bar{\vh} - \vmu_0) (\bar{\vh} - \vmu_0)^\top \nonumber
\end{align}
where $\bar{\vh}$ is the empirical mean of $\calD^{(v)}$ 
and $S$ is the scatter matrix
\begin{equation}
  S = \sum_{i=1}^{\Nv} (\vhi{i} - \bar{\vh}) (\vhi{i} - \bar{\vh})^\top
\end{equation}}\lucas{Murphy Section 4.6.3.3}

\section{Inverse-Wishart Distribution}
\label{sec:inverse-wishart}

The inverse-Wishart distribution is defined as~\citep{murphyConjugateBayesianAnalysis2007}

\mathcheck{%
\begin{align}
\IW(\Sigma \mid &\Lambda^{-1}, \nu) = \frac{1}{Z} |\Sigma|^{-\frac{\nu + d + 1}{2}} \nonumber \\
 &\times \exp \left( -\frac{1}{2}\Tr (\Lambda \Sigma^{-1}) \right)
\end{align}
where
\begin{equation}
    Z = \frac{|\Lambda|^{\frac{\nu}{2}}}{2^{\frac{\nu d}{2}} \Gamma_d(\frac{\nu}{2})}
\end{equation}
where $\Sigma$ is a positive-definite $d \times d$ matrix, and $\Gamma_d$ is the multivariate Gamma function.}\lucas{Check Murphy 2007 (not the textbook) 10.5 for this formulation of the IW.}

\section{Changes to UD Annotations}
\label{sec:ud-changes}

We apply some post-processing to canonicalize the automatically-converted \unimorph annotations.
The changes we make are:
\begin{enumerate}
    \item We remove any annotations with disjunctions. These constitute a minority of annotations, and handling them adequately requires language-specific knowledge.
    \item We fix some annotations that we believe are typos, e.g.\, replace ``\{CMPR\}'' with ``CMPR''.
    \item We let ``PST+PRF'' be a Tense annotation. This is a recurrent annotation in Latin, Romanian and Turkish.
    \item We canonicalize conjunctive features by sorting them alphabetically, ensuring they all belong to the same morphological attribute, and joining them into a new feature. So the annotation ``MASC+FEM'' becomes ``FEM+MASC''.
    \item We discard language-specific annotations as this is not a canonical \unimorph dimension.
\end{enumerate}

\section{Mutual Information Approximation}
\label{sec:mutual-information-approx}
Let $V_a$ be a $\calVa$-valued random variable denoting the value of a morphosyntactic attribute, and $H$ be a $\R^d$-valued random variable for the word representation.
The mutual information between $V_a$ and $H$ is
\begin{equation}
\MI(V_a ; H) = \ent(V_a) - \ent(V_a \mid H)
\end{equation}

To compute the entropy \mathcheck{$\ent(V_a)$}, we would ideally need the true attribute distribution \mathcheck{$\truep(v)$} for a language.
We can empirically approximate it using \mathcheck{$p(v)$}, which has been computed from held-out data
\mathcheck{%
\begin{align}
   \ent(V_a) &= \sum_{v \in \calV_a} \truep(v) \log_2 \frac{1}{\truep(v)} \\
    & \approx \sum_{v \in \calV_a} p(v) \log_2 \frac{1}{p(v)}
\end{align}}

Computing \mathcheck{$\ent(V_a \mid H)$} is trickier
as it relies on the true distribution of the representations for a
value, \mathcheck{$\truep(\vh \mid v)$}, which is hard to estimate
as it is high-dimensional and poorly sampled in our data.
\mathcheck{%
\begin{align}
    &\ent(V_a \mid H) \label{eq:conditional-entropy} \\
    &= \int  \sum_{v \in \calV_a} \truep(v,\vh) \log_2 \frac{1}{\truep(v \mid \vh)}\,\mathrm{d}\vh \nonumber \\
    &= \sum_{v \in \calV_a} \truep(v) \int \truep(\vh \mid v) \log_2 \frac{1}{\truep(v \mid \vh)}\,\mathrm{d} \vh \nonumber
\end{align}}
Note that by using an approximation \mathcheck{$\truep(v \mid \vh) \approx p(v \mid \vh)$}
instead (a.k.a.\, our probe),
\mathcheck{%
we obtain an upper bound on the true conditional
entropy~\citep{brownEstimateUpperBound1992}
\begin{align}
  &\ent(V_a \mid H) \leq \ent_{p} (V_a \mid H) \\
  &= \sum_{v \in \calV_a} \truep(v) \underbrace{
    \int \truep(\vh \mid v) \log_2 \frac{1}{p(v \mid \vh)}\,\mathrm{d} \vh
    }_{I_v} \nonumber
\end{align}}

While \mathcheck{$\truep(v) \approx p(v)$} should be reasonable for our purposes, the integral \mathcheck{$I_v$} is intractable as it still depends on \mathcheck{$\truep(\vh \mid v)$}.
However, \mathcheck{we can use held-out data to approximate $I_v$~\citep{pimentelMeaningFormMeasuring2019}
\begin{align}
  I_v &= -\int \truep(\vh \mid v) \log_2 p(v \mid \vh) \,\mathrm{d} \vh \\
    & \approx -\frac{1}{\Nv} \sum_{i=1}^{\Nv} \log_2 p(v \mid \tilde{\vh}^{(i)})
\end{align}}
where \mathcheck{$\{ {\tilde{\vh}}^{(i)} \}_{i=1}^{\Nv}$} are held-out word representations for a value \mathcheck{$v$}, and \mathcheck{thus obtain an empirical upper-bound on $\ent (V_a \mid H)$}.

\section{Reproducibility Details}

All experiments were run on an AWS \texttt{p2.xlarge} instance, with 1 Tesla K80 GPU, 4 CPU cores, and 61 GB of RAM.
The total runtime of the experiments was 2 days, 18 hours, 42 minutes and 14 seconds.

In total, when considering a \mathcheck{$d$}-dimensional word representation, this model has
\mathcheck{%
\begin{equation}
\underbrace{|\calVa| \left( \frac{d (d+1)}{2} + d \right)}_\text{Gaussians} + \underbrace{\left(|\calVa| - 1\right)}_\text{Categorical}
\end{equation}}
parameters.
In practice, this means that for every value, a \fasttext Gaussian we fit has \mathcheck{$45450$} parameters, whereas a \bert Gaussian has \mathcheck{$296064$} parameters.

\section{Probed Attributes by Language}
\label{sec:probed-attributes}

\cref{tab:probed-attributes} shows a list of all languages that were probed, which attributes were probed, and which values were considered.
The number of example words for a value in the train/validation/test split is shown in parenthesis.

\clearpage
\onecolumn
\begin{tabularx}{\textwidth}{@{}llX@{}}
\caption{Table of attributes that were probed for each language, and the values that were considered for that attribute. The number of example words for a value in the train/validation/test split is shown in parenthesis.}
\label{tab:probed-attributes}\\
    \toprule
    \textbf{Language} & \textbf{Attribute} & \textbf{Values} \\
    \midrule
    \endfirsthead
    \toprule
    \textbf{Language} & \textbf{Attribute} & \textbf{Values} \\
    \midrule
    \endhead
    \midrule
    \multicolumn{3}{r}{\footnotesize(Continued next page)}
    \endfoot
    \bottomrule
    \endlastfoot
    afr (Afrikaans) & Number & PL (2682{\breakslash}399{\breakslash}1067), SG (6390{\breakslash}999{\breakslash}1656) \\ \midrule
ara (Arabic) & Number & PL (18193{\breakslash}2282{\breakslash}2411), SG (97436{\breakslash}12692{\breakslash}12451) \\
  & Gender and Noun Class & FEM (22104{\breakslash}2666{\breakslash}2842), MASC (27953{\breakslash}3982{\breakslash}3639) \\
  & Mood & IND (6452{\breakslash}832{\breakslash}774), SBJV (1021{\breakslash}157{\breakslash}135) \\
  & Aspect & IPFV (7986{\breakslash}1050{\breakslash}999), PFV (8951{\breakslash}1292{\breakslash}1226) \\
  & Voice & ACT (16039{\breakslash}2169{\breakslash}2081), PASS (898{\breakslash}173{\breakslash}144) \\
  & Case & ACC (21975{\breakslash}2951{\breakslash}2857), GEN (70767{\breakslash}8920{\breakslash}9137), NOM (13901{\breakslash}1859{\breakslash}1668) \\
  & Definiteness & DEF (47204{\breakslash}5785{\breakslash}6077), INDF (21122{\breakslash}3004{\breakslash}2668) \\ \midrule
bel (Belarusian) & Case & GEN (912{\breakslash}336{\breakslash}262), NOM (673{\breakslash}174{\breakslash}171) \\
  & Gender and Noun Class & FEM (910{\breakslash}270{\breakslash}194), MASC (1059{\breakslash}344{\breakslash}351) \\
  & Number & PL (781{\breakslash}259{\breakslash}212), SG (2208{\breakslash}639{\breakslash}615) \\ \midrule
bul (Bulgarian) & Gender and Noun Class & FEM (16442{\breakslash}2142{\breakslash}2119), MASC (21236{\breakslash}2614{\breakslash}2650), NEUT (9292{\breakslash}1271{\breakslash}1214) \\
  & Number & PL (18973{\breakslash}2443{\breakslash}2371), SG (49940{\breakslash}6427{\breakslash}6388) \\
  & Definiteness & DEF (15310{\breakslash}1939{\breakslash}1942), INDF (33516{\breakslash}4340{\breakslash}4232) \\
  & Tense & PRS (10781{\breakslash}1405{\breakslash}1330), PST (5373{\breakslash}677{\breakslash}716) \\
  & Person & 1 (2548{\breakslash}353{\breakslash}345), 3 (14882{\breakslash}1885{\breakslash}1824) \\
  & Voice & ACT (1885{\breakslash}239{\breakslash}222), PASS (1625{\breakslash}221{\breakslash}204) \\ \midrule
cat (Catalan) & Gender and Noun Class & FEM (66961{\breakslash}9409{\breakslash}9368), MASC (85011{\breakslash}11313{\breakslash}11473) \\
  & Number & PL (54636{\breakslash}7105{\breakslash}7314), SG (150183{\breakslash}20733{\breakslash}20682) \\
  & Mood & IND (27555{\breakslash}3678{\breakslash}3662), SBJV (2070{\breakslash}303{\breakslash}252) \\
  & Tense & FUT (3005{\breakslash}319{\breakslash}405), PRS (25110{\breakslash}3347{\breakslash}3347), PST (8398{\breakslash}1236{\breakslash}1040) \\ \midrule
ces (Czech) & Case & ACC (140691{\breakslash}19275{\breakslash}20747), DAT (31793{\breakslash}4458{\breakslash}4605), ESS (104763{\breakslash}14467{\breakslash}15519), GEN (176912{\breakslash}23678{\breakslash}25261), INS (53879{\breakslash}7312{\breakslash}8282), NOM (158994{\breakslash}21358{\breakslash}23042) \\
  & Gender and Noun Class & FEM (88003{\breakslash}11924{\breakslash}13173), MASC (137896{\breakslash}18345{\breakslash}19153), NEUT (44566{\breakslash}6295{\breakslash}6682) \\
  & Comparison & CMPR (6134{\breakslash}826{\breakslash}908), RL (3199{\breakslash}442{\breakslash}494) \\
  & Number & PL (180092{\breakslash}24725{\breakslash}26325), SG (459202{\breakslash}62398{\breakslash}67686) \\
  & Person & 1 (12691{\breakslash}1993{\breakslash}2293), 2 (1973{\breakslash}342{\breakslash}471), 3 (68973{\breakslash}9461{\breakslash}10390) \\
  & Aspect & IPFV (41460{\breakslash}5706{\breakslash}6268), PFV (29944{\breakslash}4151{\breakslash}4408) \\
  & Tense & PRS (64246{\breakslash}8849{\breakslash}10059), PST (44390{\breakslash}6089{\breakslash}6523) \\
  & Polarity & NEG (16126{\breakslash}2172{\breakslash}2361), POS (86217{\breakslash}11270{\breakslash}12221) \\
  & Animacy & ANIM (55084{\breakslash}7179{\breakslash}7543), INAN (62155{\breakslash}8469{\breakslash}8955) \\
  & Voice & ACT (3549{\breakslash}410{\breakslash}539), PASS (7426{\breakslash}1044{\breakslash}1056) \\ \midrule
dan (Danish) & Gender and Noun Class & FEM+MASC (16075{\breakslash}2045{\breakslash}1981), NEUT (7294{\breakslash}964{\breakslash}872) \\
  & Definiteness & DEF (5218{\breakslash}664{\breakslash}655), INDF (14149{\breakslash}1867{\breakslash}1682) \\
  & Number & PL (7332{\breakslash}1050{\breakslash}909), SG (21782{\breakslash}2784{\breakslash}2639) \\
  & Tense & PRS (5806{\breakslash}753{\breakslash}679), PST (4017{\breakslash}575{\breakslash}604) \\ \midrule
deu (German) & Number & PL (17392{\breakslash}1009{\breakslash}1259), SG (78706{\breakslash}3789{\breakslash}4698) \\
  & Case & ACC (20352{\breakslash}1243{\breakslash}1480), DAT (29961{\breakslash}1150{\breakslash}1694), GEN (5675{\breakslash}195{\breakslash}314), NOM (28192{\breakslash}1528{\breakslash}1729) \\ \midrule
eng (English) & Number & PL (12599{\breakslash}1376{\breakslash}1364), SG (55978{\breakslash}7192{\breakslash}7266) \\
  & Tense & PRS (8129{\breakslash}1063{\breakslash}940), PST (9359{\breakslash}996{\breakslash}981) \\ \midrule
est (Estonian) & Case & ABL+IN (1383{\breakslash}155{\breakslash}169), ALL+AT (1299{\breakslash}154{\breakslash}175), ALL+IN (1451{\breakslash}166{\breakslash}188), AT+ESS (1813{\breakslash}241{\breakslash}221), COM (1011{\breakslash}131{\breakslash}129), ESS+IN (1757{\breakslash}210{\breakslash}215), GEN (8808{\breakslash}1081{\breakslash}1132), NOM (13955{\breakslash}1727{\breakslash}1683), PRT (5022{\breakslash}572{\breakslash}628) \\
  & Number & PL (8434{\breakslash}1052{\breakslash}1001), SG (38059{\breakslash}4655{\breakslash}4801) \\
  & Finiteness & FIN (11753{\breakslash}1462{\breakslash}1501), NFIN (1306{\breakslash}181{\breakslash}170) \\
  & Tense & PRS (5633{\breakslash}670{\breakslash}680), PST (6734{\breakslash}894{\breakslash}856) \\
  & Person & 1 (2240{\breakslash}252{\breakslash}312), 3 (9058{\breakslash}1175{\breakslash}1144) \\ \midrule
eus (Basque) & Case & ABL+AT (532{\breakslash}163{\breakslash}187), ABS (10459{\breakslash}3457{\breakslash}3465), ALL+AT (514{\breakslash}181{\breakslash}169), COM (383{\breakslash}128{\breakslash}148), DAT (745{\breakslash}232{\breakslash}239), ERG (2670{\breakslash}859{\breakslash}873), ESS (3148{\breakslash}977{\breakslash}1024), ESS+IN (3408{\breakslash}1167{\breakslash}1180), GEN (2334{\breakslash}763{\breakslash}806), INS (633{\breakslash}235{\breakslash}203), PRT (420{\breakslash}135{\breakslash}162) \\
  & Animacy & ANIM (778{\breakslash}274{\breakslash}236), INAN (7269{\breakslash}2375{\breakslash}2521) \\
  & Definiteness & DEF (19134{\breakslash}6315{\breakslash}6336), INDF (3688{\breakslash}1244{\breakslash}1224) \\
  & Number & PL (4162{\breakslash}1393{\breakslash}1376), SG (15257{\breakslash}5017{\breakslash}5057) \\
  & Aspect & IPFV (1062{\breakslash}363{\breakslash}395), PFV (3476{\breakslash}1149{\breakslash}1140), PROG (2937{\breakslash}914{\breakslash}967), PROSP (953{\breakslash}297{\breakslash}279) \\ \midrule
fas (Persian) & Number & PL (11152{\breakslash}1250{\breakslash}1327), SG (50635{\breakslash}7040{\breakslash}7105) \\ \midrule
fin (Finnish) & Number & PL (21315{\breakslash}2356{\breakslash}2878), SG (79259{\breakslash}8978{\breakslash}9967) \\
  & Case & ABL+IN (4204{\breakslash}487{\breakslash}531), ALL+AT (1909{\breakslash}236{\breakslash}254), ALL+IN (5014{\breakslash}539{\breakslash}616), AT+ESS (3310{\breakslash}375{\breakslash}384), ESS+IN (5508{\breakslash}600{\breakslash}661), FRML (1974{\breakslash}214{\breakslash}261), GEN (20002{\breakslash}2299{\breakslash}2490), NOM (25818{\breakslash}2905{\breakslash}3252), PRT (12638{\breakslash}1404{\breakslash}1709), TRANS (1206{\breakslash}111{\breakslash}139) \\
  & Voice & ACT (23469{\breakslash}2626{\breakslash}3082), PASS (4179{\breakslash}505{\breakslash}542) \\
  & Tense & PRS (11149{\breakslash}1314{\breakslash}1732), PST (9039{\breakslash}980{\breakslash}958) \\
  & Person & 1 (3104{\breakslash}363{\breakslash}412), 3 (15218{\breakslash}1746{\breakslash}2007) \\ \midrule
fra (French) & Gender and Noun Class & FEM (63408{\breakslash}6471{\breakslash}1623), MASC (81523{\breakslash}8352{\breakslash}2439) \\
  & Number & PL (41157{\breakslash}4146{\breakslash}1286), SG (131994{\breakslash}13416{\breakslash}3681) \\
  & Tense & PRS (19256{\breakslash}1864{\breakslash}589), PST (14020{\breakslash}1382{\breakslash}343) \\ \midrule
gle (Irish) & Gender and Noun Class & FEM (327{\breakslash}1240{\breakslash}1158), MASC (690{\breakslash}2188{\breakslash}2208) \\
  & Number & PL (177{\breakslash}752{\breakslash}594), SG (1181{\breakslash}3841{\breakslash}3841) \\ \midrule
heb (Hebrew) & Definiteness & DEF (2184{\breakslash}174{\breakslash}156), INDF (21817{\breakslash}1812{\breakslash}2069) \\
  & Number & PL (14478{\breakslash}1328{\breakslash}1280), SG (38263{\breakslash}3182{\breakslash}3650) \\ \midrule
hin (Hindi) & Number & PL (24553{\breakslash}3049{\breakslash}2932), SG (149419{\breakslash}18658{\breakslash}19128) \\
  & Case & ACC (79132{\breakslash}9903{\breakslash}10138), NOM (66735{\breakslash}8392{\breakslash}8437) \\
  & Gender and Noun Class & FEM (43951{\breakslash}5496{\breakslash}5686), MASC (104389{\breakslash}13116{\breakslash}13253) \\ \midrule
hrv (Croatian) & Gender and Noun Class & FEM (31053{\breakslash}3094{\breakslash}2468), MASC (41905{\breakslash}3285{\breakslash}3084), NEUT (12411{\breakslash}921{\breakslash}1070) \\
  & Number & PL (27716{\breakslash}2672{\breakslash}2583), SG (74308{\breakslash}5976{\breakslash}5357) \\
  & Case & ACC (22562{\breakslash}2038{\breakslash}1721), DAT (2332{\breakslash}197{\breakslash}171), ESS (14876{\breakslash}1335{\breakslash}1182), GEN (27281{\breakslash}2552{\breakslash}2163), INS (5388{\breakslash}366{\breakslash}398), NOM (26435{\breakslash}2125{\breakslash}2046) \\
  & Tense & PRS (15665{\breakslash}1298{\breakslash}1299), PST (6537{\breakslash}509{\breakslash}436) \\
  & Finiteness & FIN (16468{\breakslash}1349{\breakslash}1326), NFIN (3331{\breakslash}288{\breakslash}273) \\ \midrule
hun (Hungarian) & Definiteness & DEF (2885{\breakslash}1770{\breakslash}1524), INDF (1307{\breakslash}577{\breakslash}619) \\
  & Number & PL (1516{\breakslash}850{\breakslash}744), SG (9948{\breakslash}5853{\breakslash}5223) \\
  & Case & ACC (935{\breakslash}541{\breakslash}484), ALL+ON (248{\breakslash}162{\breakslash}157), ESS+IN (478{\breakslash}248{\breakslash}242), INS (218{\breakslash}198{\breakslash}155), NOM (6492{\breakslash}3910{\breakslash}3352) \\
  & Possession & PSS3S (1139{\breakslash}775{\breakslash}652), PSSD (5676{\breakslash}2779{\breakslash}2353) \\
  & Tense & PRS (956{\breakslash}513{\breakslash}369), PST (795{\breakslash}357{\breakslash}533) \\ \midrule
ita (Italian) & Gender and Noun Class & FEM (44923{\breakslash}1947{\breakslash}1713), MASC (59063{\breakslash}2613{\breakslash}2265) \\
  & Number & PL (38689{\breakslash}1739{\breakslash}1321), SG (95035{\breakslash}4138{\breakslash}3843) \\
  & Tense & PRS (15854{\breakslash}693{\breakslash}620), PST (11200{\breakslash}491{\breakslash}432) \\ \midrule
lat (Latin) & Number & PL (1237{\breakslash}2086{\breakslash}1757), SG (3726{\breakslash}4029{\breakslash}4883) \\
  & Case & ABL+AT (944{\breakslash}1150{\breakslash}999), ACC (1369{\breakslash}1545{\breakslash}1813), DAT (231{\breakslash}306{\breakslash}270), GEN (492{\breakslash}451{\breakslash}324), NOM (809{\breakslash}1353{\breakslash}1436) \\
  & Gender and Noun Class & FEM (517{\breakslash}721{\breakslash}621), MASC (912{\breakslash}1187{\breakslash}1150), NEUT (378{\breakslash}570{\breakslash}525) \\
  & Person & 1 (179{\breakslash}166{\breakslash}324), 3 (837{\breakslash}892{\breakslash}1232) \\
  & Tense & PRS (694{\breakslash}1020{\breakslash}1224), PST (815{\breakslash}764{\breakslash}1047) \\
  & Mood & IND (868{\breakslash}939{\breakslash}1431), SBJV (212{\breakslash}255{\breakslash}270) \\
  & Aspect & IPFV (138{\breakslash}209{\breakslash}246), PFV (689{\breakslash}567{\breakslash}814) \\ \midrule
lav (Latvian) & Case & ACC (5729{\breakslash}1113{\breakslash}1139), DAT (2999{\breakslash}622{\breakslash}610), ESS (3148{\breakslash}619{\breakslash}704), GEN (7251{\breakslash}1343{\breakslash}1311), NOM (10222{\breakslash}2257{\breakslash}2300) \\
  & Number & PL (8157{\breakslash}1494{\breakslash}1678), SG (21128{\breakslash}4474{\breakslash}4517) \\
  & Gender and Noun Class & FEM (5243{\breakslash}987{\breakslash}1029), MASC (6252{\breakslash}1276{\breakslash}1319) \\
  & Tense & PRS (4629{\breakslash}838{\breakslash}1129), PST (3673{\breakslash}1015{\breakslash}749) \\
  & Person & 1 (1539{\breakslash}436{\breakslash}450), 3 (6449{\breakslash}1368{\breakslash}1332) \\ \midrule
lit (Lithuanian) & Case & GEN (356{\breakslash}153{\breakslash}150), NOM (504{\breakslash}164{\breakslash}152) \\
  & Gender and Noun Class & FEM (496{\breakslash}162{\breakslash}159), MASC (805{\breakslash}282{\breakslash}296) \\
  & Number & PL (459{\breakslash}180{\breakslash}215), SG (1176{\breakslash}373{\breakslash}342) \\ \midrule
nld (Dutch) & Number & PL (10797{\breakslash}615{\breakslash}793), SG (42640{\breakslash}2850{\breakslash}2609) \\
  & Finiteness & FIN (17418{\breakslash}1023{\breakslash}903), NFIN (5213{\breakslash}242{\breakslash}407) \\
  & Gender and Noun Class & FEM+MASC (18298{\breakslash}1316{\breakslash}1225), NEUT (10238{\breakslash}687{\breakslash}690) \\ \midrule
pol (Polish) & Case & ACC (7083{\breakslash}1188{\breakslash}1278), ESS (5790{\breakslash}859{\breakslash}876), GEN (10429{\breakslash}1663{\breakslash}1773), INS (2616{\breakslash}502{\breakslash}463), NOM (7575{\breakslash}1228{\breakslash}1268) \\
  & Number & PL (9871{\breakslash}1491{\breakslash}1573), SG (25225{\breakslash}4242{\breakslash}4454) \\
  & Gender and Noun Class & FEM (4083{\breakslash}678{\breakslash}755), MASC (7858{\breakslash}1306{\breakslash}1380), NEUT (2416{\breakslash}371{\breakslash}409) \\
  & Animacy & HUM (4285{\breakslash}663{\breakslash}755), INAN (1641{\breakslash}268{\breakslash}309) \\
  & Tense & PRS (3823{\breakslash}634{\breakslash}660), PST (3547{\breakslash}621{\breakslash}645) \\
  & Person & 1 (1501{\breakslash}261{\breakslash}332), 3 (3725{\breakslash}613{\breakslash}603) \\ \midrule
por (Portuguese) & Number & PL (27002{\breakslash}1366{\breakslash}1299), SG (92097{\breakslash}5125{\breakslash}4723) \\
  & Gender and Noun Class & FEM (40907{\breakslash}2138{\breakslash}2107), MASC (57079{\breakslash}3154{\breakslash}2850) \\
  & Tense & PRS (8438{\breakslash}512{\breakslash}466), PST (9107{\breakslash}470{\breakslash}449) \\ \midrule
ron (Romanian) & Definiteness & DEF (24561{\breakslash}2326{\breakslash}2199), INDF (33780{\breakslash}3142{\breakslash}2992) \\
  & Number & PL (28550{\breakslash}2558{\breakslash}2430), SG (66435{\breakslash}6248{\breakslash}6013) \\
  & Mood & IND (11099{\breakslash}1000{\breakslash}975), SBJV (3623{\breakslash}390{\breakslash}329) \\
  & Gender and Noun Class & FEM (17544{\breakslash}1687{\breakslash}1510), MASC (14229{\breakslash}1315{\breakslash}1333) \\ \midrule
rus (Russian) & Animacy & ANIM (7032{\breakslash}1184{\breakslash}1156), INAN (32548{\breakslash}5037{\breakslash}4869) \\
  & Case & ACC (5262{\breakslash}831{\breakslash}834), DAT (1732{\breakslash}207{\breakslash}248), ESS (5066{\breakslash}751{\breakslash}807), GEN (13687{\breakslash}2201{\breakslash}2089), INS (3041{\breakslash}452{\breakslash}428), NOM (12342{\breakslash}2017{\breakslash}1831) \\
  & Gender and Noun Class & FEM (11145{\breakslash}1842{\breakslash}1762), MASC (21073{\breakslash}3360{\breakslash}3309), NEUT (6774{\breakslash}961{\breakslash}953) \\
  & Number & PL (9691{\breakslash}1432{\breakslash}1413), SG (34647{\breakslash}5518{\breakslash}5385) \\
  & Tense & PRS (1870{\breakslash}293{\breakslash}275), PST (4227{\breakslash}631{\breakslash}677) \\
  & Aspect & IPFV (3978{\breakslash}602{\breakslash}619), PFV (3133{\breakslash}481{\breakslash}498) \\
  & Voice & MID (1326{\breakslash}192{\breakslash}208), PASS (1125{\breakslash}178{\breakslash}173) \\ \midrule
slk (Slovak) & Gender and Noun Class & FEM (14217{\breakslash}2249{\breakslash}2566), MASC (17129{\breakslash}3838{\breakslash}3450), NEUT (6817{\breakslash}992{\breakslash}1306) \\
  & Number & PL (8989{\breakslash}1635{\breakslash}2013), SG (36266{\breakslash}5750{\breakslash}5840) \\
  & Case & ACC (9651{\breakslash}1392{\breakslash}1466), DAT (2031{\breakslash}328{\breakslash}271), ESS (5062{\breakslash}1203{\breakslash}1151), GEN (7228{\breakslash}1867{\breakslash}1998), INS (3108{\breakslash}699{\breakslash}698), NOM (10605{\breakslash}1869{\breakslash}2131) \\
  & Tense & PRS (4926{\breakslash}282{\breakslash}491), PST (8271{\breakslash}950{\breakslash}823) \\
  & Aspect & IPFV (8561{\breakslash}705{\breakslash}898), PFV (6003{\breakslash}608{\breakslash}524) \\
  & Animacy & ANIM (8769{\breakslash}2069{\breakslash}1401), INAN (8360{\breakslash}1769{\breakslash}2049) \\ \midrule
slv (Slovenian) & Case & ACC (13762{\breakslash}1794{\breakslash}1709), DAT (2219{\breakslash}236{\breakslash}257), ESS (10546{\breakslash}1448{\breakslash}1273), GEN (12424{\breakslash}1667{\breakslash}1545), INS (4713{\breakslash}673{\breakslash}630), NOM (13405{\breakslash}1615{\breakslash}1730) \\
  & Gender and Noun Class & FEM (9549{\breakslash}1149{\breakslash}1231), MASC (13512{\breakslash}1642{\breakslash}1626), NEUT (4732{\breakslash}597{\breakslash}610) \\
  & Number & PL (18042{\breakslash}2692{\breakslash}2286), SG (44944{\breakslash}5221{\breakslash}5650) \\
  & Person & 1 (2120{\breakslash}275{\breakslash}253), 3 (11322{\breakslash}1247{\breakslash}1485) \\
  & Finiteness & FIN (12361{\breakslash}1474{\breakslash}1568), NFIN (1083{\breakslash}163{\breakslash}146) \\
  & Aspect & IPFV (4774{\breakslash}580{\breakslash}649), PFV (5233{\breakslash}602{\breakslash}623) \\ \midrule
spa (Spanish) & Number & PL (47382{\breakslash}4347{\breakslash}1471), SG (139165{\breakslash}13604{\breakslash}4494) \\
  & Gender and Noun Class & FEM (60665{\breakslash}5724{\breakslash}1857), MASC (79816{\breakslash}7849{\breakslash}2522) \\
  & Tense & PRS (16120{\breakslash}1520{\breakslash}644), PST (13814{\breakslash}1336{\breakslash}381) \\ \midrule
srp (Serbian) & Number & PL (10057{\breakslash}1606{\breakslash}1754), SG (31875{\breakslash}4781{\breakslash}5141) \\
  & Gender and Noun Class & FEM (12944{\breakslash}1928{\breakslash}2193), MASC (17331{\breakslash}2597{\breakslash}2793), NEUT (4187{\breakslash}621{\breakslash}626) \\
  & Case & ACC (8294{\breakslash}1329{\breakslash}1407), DAT (866{\breakslash}126{\breakslash}163), ESS (5804{\breakslash}882{\breakslash}1038), GEN (10910{\breakslash}1547{\breakslash}1745), INS (2029{\breakslash}262{\breakslash}261), NOM (11456{\breakslash}1748{\breakslash}1816) \\ \midrule
swe (Swedish) & Gender and Noun Class & FEM+MASC (4813{\breakslash}757{\breakslash}1403), NEUT (2730{\breakslash}457{\breakslash}840) \\
  & Number & PL (8110{\breakslash}1254{\breakslash}2721), SG (18229{\breakslash}2638{\breakslash}5248) \\
  & Definiteness & DEF (10447{\breakslash}1775{\breakslash}3192), INDF (17005{\breakslash}2321{\breakslash}5116) \\ \midrule
tur (Turkish) & Case & ABL+AT (709{\breakslash}175{\breakslash}183), ACC (1688{\breakslash}428{\breakslash}451), DAT (1837{\breakslash}436{\breakslash}489), ESS (1415{\breakslash}361{\breakslash}359), GEN (1540{\breakslash}380{\breakslash}385), INS (515{\breakslash}139{\breakslash}123), NOM (8690{\breakslash}2288{\breakslash}2362) \\
  & Aspect & IPFV (722{\breakslash}232{\breakslash}214), PFV (6156{\breakslash}1589{\breakslash}1671), PROG (887{\breakslash}248{\breakslash}225) \\
  & Person & 1 (1433{\breakslash}392{\breakslash}348), 2 (624{\breakslash}189{\breakslash}147), 3 (7013{\breakslash}1867{\breakslash}1880) \\
  & Tense & PRS (3563{\breakslash}945{\breakslash}963), PST (2941{\breakslash}733{\breakslash}757) \\
  & Number & PL (2737{\breakslash}687{\breakslash}729), SG (16222{\breakslash}4262{\breakslash}4283) \\
  & Possession & PSS1S (531{\breakslash}126{\breakslash}141), PSS3S (4035{\breakslash}982{\breakslash}1053) \\
  & Polarity & NEG (782{\breakslash}227{\breakslash}237), POS (6410{\breakslash}1694{\breakslash}1713) \\ \midrule
ukr (Ukrainian) & Case & ACC (8908{\breakslash}1196{\breakslash}1681), ESS (4895{\breakslash}656{\breakslash}997), GEN (12499{\breakslash}2087{\breakslash}3397), INS (3953{\breakslash}505{\breakslash}843), NOM (9919{\breakslash}1398{\breakslash}1831) \\
  & Number & PL (11432{\breakslash}1507{\breakslash}2215), SG (28210{\breakslash}4031{\breakslash}6016) \\
  & Gender and Noun Class & FEM (4716{\breakslash}556{\breakslash}1035), MASC (6245{\breakslash}890{\breakslash}1294), NEUT (2600{\breakslash}318{\breakslash}477) \\
  & Animacy & ANIM (2671{\breakslash}316{\breakslash}447), INAN (2696{\breakslash}407{\breakslash}615) \\
  & Tense & PRS (2505{\breakslash}397{\breakslash}454), PST (4093{\breakslash}380{\breakslash}535) \\ \midrule
urd (Urdu) & Number & PL (8105{\breakslash}1008{\breakslash}844), SG (58067{\breakslash}7841{\breakslash}8254) \\
  & Case & ACC (29707{\breakslash}4210{\breakslash}4112), NOM (29217{\breakslash}3853{\breakslash}4264) \\
\end{tabularx}
\clearpage
\twocolumn

\end{document}